%% file: main.tex
\documentclass[nopreprintline,12pt]{elsarticle}
\usepackage[utf8]{inputenc}

\usepackage{bm}
\usepackage{mystyle}
\usepackage{appendix}
\usepackage{geometry}
\usepackage{wrapfig}

\usepackage{pgfplots}
\pgfplotsset{compat=1.9}
\pgfmathdeclarefunction{gauss}{2}{%
  \pgfmathparse{1/(#2*sqrt(2*pi))*exp(-((x-#1)^2)/(2*#2^2))}%
}
\pgfmathdeclarefunction{sigmoid}{2}{%
  \pgfmathparse{1/(1+exp(-(x-#1)/#2))}%
}

\usetikzlibrary{arrows.meta}
\usetikzlibrary{decorations.pathreplacing,decorations.markings}

\tikzset{T/.tip={Triangle[length=2mm, width=2mm]}}
\tikzset{L/.style = {postaction={decorate,decoration={
        markings,
        mark=at position .5 with {\arrow[xshift=0.9mm,black]{Square}}
        }},
      shorten > = 0.75mm, shorten < = 0.75mm, thick}
     }

\graphicspath{{figs/}}

\journal{Neural Networks}

\begin{document}

\begin{frontmatter}

\title{Flexibly Regularized Mixture Models and Application to Image Segmentation}

\author[cl,jv]{Jonathan Vacher}
\ead{jonathan.vacher@ens.psl.eu}
\ead[url]{https://jonathanvacher.github.io/}
\author[cl]{Claire Launay}
\affiliation[cl]{organization={Dept. of Systems and Computational Biology, Albert Einstein College of Medicine},
            addressline={1300 Morris Park Ave},
            city={Bronx},
            postcode={10461},
            state={NY},
            country={USA}}
\ead{claire.launay@einsteinmed.edu}
\author[cl,rcc,rcc2]{Ruben Coen-Cagli}
\affiliation[rcc]{organization={Dominick P. Purpura Dept. of Neuroscience, Albert Einstein College of Medicine},
            addressline={1300 Morris Park Ave},
            city={Bronx},
            postcode={10461},
            state={NY},
            country={USA}}
\affiliation[rcc2]{organization={Dept. of Ophthalmology \& Visual Sciences, Albert Einstein College of Medicine},
            addressline={1300 Morris Park Ave},
            city={Bronx},
            postcode={10461},
            state={NY},
            country={USA}}
\ead{ruben.coen-cagli@einsteinmed.edu}
\affiliation[jv]{organization={Laboratoire des Systèmes Perceptif, Département d'\'Etudes Cognitives, \'Ecole Normale Supérieure, PSL University},
            addressline={24 rue Lhomond, Bâtiment Jaurès, 2éme étage},
            city={Paris},
            postcode={75005},
            state={},
            country={France}}

\begin{abstract}
Probabilistic finite mixture models are widely used for unsupervised clustering. These models can often be improved by adapting them to the topology of the data. For instance, in order to classify spatially adjacent data points similarly, it is common to introduce a Laplacian constraint on the posterior probability that each data point belongs to a class. Alternatively, the mixing probabilities can be treated as free parameters, while assuming Gauss-Markov or more complex priors to regularize those mixing probabilities. However, these approaches are constrained by the shape of the prior and often lead to complicated or intractable inference. Here, we propose a new parametrization of the Dirichlet distribution to flexibly regularize the mixing probabilities of over-parametrized mixture distributions. Using the Expectation-Maximization algorithm, we show that our approach allows us to define any linear update rule for the mixing probabilities, including spatial smoothing regularization as a special case. We then show that this flexible design can be extended to share class information between multiple mixture models. We apply our algorithm to artificial and natural image segmentation tasks, and we provide quantitative and qualitative comparison of the performance of Gaussian and Student-t mixtures on the Berkeley Segmentation Dataset. We also demonstrate how to propagate class information across the layers of deep convolutional neural networks in a probabilistically optimal way, suggesting a new interpretation for feedback signals in biological visual systems. Our flexible approach can be easily generalized to adapt probabilistic mixture models to arbitrary data topologies.
\end{abstract}




\end{frontmatter}

\section{Introduction}
\label{sec:intro}

Probabilistic finite mixture models are a class of statistical models that assume the density of observed data is a weighted sum of simpler component distributions. Finite mixture models aggregate data points by their statistical similarity and are widely applied to unsupervised clustering problems~\cite{mclachlan2019finite}. Because these models do not take into account topological dependencies between the observed data points, they often result in scattered clusters of data points. In practice, clustering results are often improved by accounting for the topology of the data~\cite{he2010laplacian,ye2018nonparametric,liu2010gaussian,gan2015manifold}. A common approach to achieve this is to introduce a dependence of the class probability of one data point on the class of the other data points, either explicitly or implicitly through structured priors that act effectively as a regularization term to the likelihood function. Here we present a new formulation of finite probabilistic mixtures that allows one to impose any topology on the class probabilities, and that offers a computationally efficient optimization of the model's parameters. To illustrate our approach, we consider unsupervised segmentation of natural images: a paradigmatic application of finite mixture models, in which accounting for the data topology is necessary to \revisedB{reduce noise and spatial discontinuity in the segments.}

Segmentation is the task of partitioning an image into multiple areas or segments, or equivalently of assigning segment labels (\ie{} mixture components, in the language of finite mixtures) to each image pixel. Recent advances in deep learning models allowed the development of successful strategies for supervised segmentation~\cite{minaee2021image, ronneberger2015u, long2015fully, badrinarayanan2017segnet, chen2018deeplab}. These approaches are trained with labeled images and are extremely effective, in part because they learn object appearance which implicitly forces those algorithms to respect the spatial topology of images, \ie{} grouping together nearby pixels. A known limitation of these supervised strategies is that they do not learn to group pixels by their similarity and therefore generalize poorly to images of unseen object categories.

\revisedB{Unsupervised learning is an abundant field gathering tree methods~\cite{shi2006unsupervised}, dictionary learning~\cite{thiagarajan2014multiple} and more~\cite{zhao2019recursive}}. In particular, many algorithms for unsupervised segmentation have been developed, using different definitions of similarity between pixels, such as k-means clustering~\cite{dhanachandra2015}, active contours models~\cite{kass1988snakes}, graph cuts~\cite{boykov2001fast} \revisedB{and fuzzy clustering \cite{yang2005unsupervised}}. Different from those algorithms, finite probabilistic mixture models have two advantages that, as we show in this paper, can be leveraged to \revisedB{develop segmentation algorithms that capture key aspects of biological vision} and also generalize to many other clustering problems. First, the mixture components represent an explicit model of the statistical distribution of the data points within each segment. Therefore, knowledge about the statistics of natural images \cite{aaponatural,olshausen1996emergence, wainwright2000scale,coencagli2009stat,coencagli2012surround,vacher2020texture} can help choosing an appropriate parametric family for the component distributions. Second, the probabilistic formulation offers not just a segmentation of the image, but also a measure of the uncertainty of the assignment of pixels to labels, which can be used to combine optimally multiple cues for segmentation. \revisedB{These properties are particularly important for modeling human perceptual segmentation}, because studies of biological visual processing have shown that neurons in the visual cortex of the brain are sensitive to the statistical regularities of natural images~\cite{aaponatural, wainwright2000scale,coencagli2009stat,coencagli2012surround}, and that human perception uses multiple segmentation cues tuned to those statistics~\cite{wagemans2012century, elder2002ecological, sigman2001common, martin2004learning, fowlkes2007local} and combines them near-optimally~\cite{saarela2012combination}.

Given the topology of images, existing probabilistic mixture models for segmentation have been extended to encourage the assignment of spatially neighboring pixels to the same mixture component. Many authors have proposed to add a penalty on the posterior class probabilities to enforce their local similarity~\cite{liu2010gaussian, gan2015manifold, he2010laplacian, ye2018nonparametric}. Other authors have proposed mixture models in which the prior mixing probabilities depend on the index of the sample (\eg{} the spatial position of the pixel). The problem with this approach is that, because the mixing probabilities are parameters that have to be learned, it increases the number of model parameters way above the number of samples. A commonly adopted solution is to strongly regularize the mixing probabilities, by accounting for the topology of the dataset. For example, in order to favor grouping of neighboring samples Blei and Frazier~\cite{blei2011distance} have developed the distance dependent Chinese restaurant process (ddCRP). The ddCRP has been further extended to perform image segmentation using a region-based hierarchical representation~\cite{ghosh2011spatial}. Similarly in a series of papers, Nikou~\etal{} have proposed to consider the mixing probabilities as Gauss-Markov random fields either directly~\cite{sfikas2007robust,nikou2007class,sfikas2008edge} or hierarchically~\cite{nikou2010bayesian}. More recently and using comparable ideas, Sun~\etal{} have introduced Location Dependent Dirichlet Processes (LDDP)~\cite{sun2017location} which use exponential of Gaussian processes in combination with Dirichlet processes to describe class distributions.

Our work builds on the tools and concepts used by Sun~\etal{}~\cite{sun2017location} and Nikou~\etal{}~\cite{nikou2010bayesian}, and extends their formulations to address two important limitations. First, those approaches do not allow any flexibility in defining how information about segment assignment is combined across pixels. Furthermore, during inference, this combination is typically nonlinear leading to computationally intensive, and sometimes unstable, optimization. Second, the use of \textit{ad-hoc} solutions for spatial regularization does not provide a clear route for extending those models to different topologies. This is important not only for other clustering problems defined, for instance, on temporal sequences or tree graphs, but also for image segmentation itself: in natural images, there is another topology associated with the hierarchy of visual features, \eg{} the hidden units at different layers of deep neural networks (DNN). It is well-known from image style-transfer applications~\cite{jing2019neural} that shallow layers convey texture information while the deep layers convey geometric and structural information. Combining that information in a systematic way could improve image segmentation. Studies on human perception have shown that humans are sensitive to segmentation cues at several levels~\cite{wagemans2012century}, that they can combine segmentation information from multiple levels near-optimally~\cite{saarela2012combination}, and that high level features, like objects, strongly affect segmentation in human observers \cite{peterson1994object,neri2017object}. Related work in computer vision has demonstrated that recurrent and feedback signals between feature levels~\cite{linsley2018learning,kim2019disentangling,kreiman2020feedback} are crucial for perceptual grouping and segmentation. The framework we introduce here extends naturally to this hierarchical structure, and allows for optimal weighting of recurrence and feedback signals.

In the remainder of Section~\ref{sec:intro}, we first briefly recap the formulation of probabilistic mixture models and the EM algorithm used for parameter optimization. Then, as they are closely related to our work, we describe the approaches of Sun~\etal{}~\cite{sun2017location} and Nikou~\etal{}~\cite{nikou2010bayesian} which account for image topology by regularizing the class probabilities. We next introduce our formulation and explain its advantages over those approaches. In Section~\ref{sec:linear-update-mixing-prob} we provide full details on our model for the class probabilities, which requires a directed graphical model with loops and a specific prior parametrization, and we state formally the theoretical results that come with it. Then, we show how our formulation can be readily extended to combine multiple mixture models through their mixing probabilities, thus accounting for hierarchical structure. In Section~\ref{sec:numerical-exp}, we illustrate the performance of our model in synthetic and natural image segmentation tasks.

\paragraph{Notations} We use the following notations. Integers $H$, $N$, $K$ and $D$ denote respectively, the number of layers, the number of samples, the number of classes and the dimension. A random variable is denoted by a capital letter $X$. The probability density function of $X$ is denoted $\pr{X}$ while $x_n$ denotes a sample. The set $\Delta^K$ represents the $K$-dimensional simplex. A bold letter (lowercase or capital) is a collection of $K$ variables $\mathbf{b}=(b_1,\dots,b_K)$. A set of $N$ samples $(x_n)_{1 \leq n \leq N}$ is shortened by $(x_n)_n$ or $x_.$. This notation also holds for random variables.

\subsection{Probabilistic mixture models and Expectation-Maximization}

A sample $x_n \in \RR^D$ is a $D$-dimensional feature vector which is a realization of a random vector $X_n$. When the distribution of $X_n$ is a weighted sum of $K>0$ distributions, we say that $X_n$ follows a mixture distribution with $K$ components. Each random variable $X_n$ is associated with a discrete latent random variable $C_n$ denoting its class among the $K$ components.

Specifically, we consider mixtures of parametric distributions, \ie{} for all $k \in \{1, \dots, K\}$, the distribution of any observed variable with class $C_n = k$ belongs to the same parametric family (Gaussian, Exponential, \dots). In the following, we write the distribution parameter (or set of parameters) associated with the class $k$ as $a_k$ and the probability of a sample $x_n$ given its class $C_n = k$ as $\mathbb{P}_{X^{(k)}}(x_n ;a_k)$.

Standard probabilistic mixture models \cite{mclachlan2019finite} assume that the latent random variables $(C_n)_n$ are i.i.d. and follow a multinomial distribution, \ie~for all $n \in \{1,\dots,N\}$, for all $k \in \{1,\dots,K\}$,
\eql{
\pr{C_n}(k) = p_k,
}
where $\mathbf{p}=(p_1,\ldots,p_K) \in \Delta^K$ is the vector of class probabilities, also called the mixing probabilities. Thus, given the mixing probabilities $\mathbf{p}$ and the collection of distribution parameters $\mathbf{a}=(a_1,\dots,a_K)$, the density function writes
\eql{\label{eq:stmixtmod}
\cpr{X}{\mathbf{P},\mathbf{A}}(x\vert \mathbf{p},\mathbf{a}) = \sum_{k=1}^K p_{k} \mathbb{P}_{X^{(k)}}(x;a_k).
}
The graphical model of those mixture models is shown in Figure~\ref{fig:graph-std-mix}.

Given samples $(x_n)_n$, the Maximum Likelihood (ML) or Maximum A Posteriori (MAP) estimates of the model parameters $\boldsymbol{\theta} = \left(\mathbf{p}, \mathbf{a}\right)$ can be found using the Expectation-Maximization (EM) algorithm. The EM algorithm is an iterative optimization method which proceeds in two steps: (i)~the expectation step (E-step) which aims at estimating the objective function knowing previous parameter estimates; (ii) the maximization step (M-step) which aims at maximizing the objective function estimated during the E-step in order to update the previous parameter estimates.

One approach to arrive at these two-steps is to consider the likelihood of the samples completed by their class $((x_n,C_n))_n$ \ie{}
\eqALl{\label{eq:log-lkl}
\ell\left(\mathbf{p},\mathbf{a}; (x_n, C_n)_{n} \right) &= \ln\left( \prod_{n=1}^{N} \cpr{X,C}{\mathbf{P},\mathbf{A}}(x_n, C_n \vert \mathbf{p}; \mathbf{a}) \right)\nonumber\\
&= \ln \left( \prod_{n=1}^{N} \prod_{k=1}^{K} \left(p_{k} \pr{X^{(k)}}(x_n; a_k) \right)^{\one_k(C_n)}\right)
}
where
\eql{
 \one_j(i) = \begin{cases} 1 & \text{if } i=j,\\ 0 & \text{otherwise}. \end{cases}
}
Such a completion turns the sum in Equation~\eqref{eq:stmixtmod} into a product which in turn will behave nicely when considering the log-likelihood. Indeed Equation~\eqref{eq:log-lkl} becomes
\eql{
\ell\left(\mathbf{p},\mathbf{a}; (x_n, C_n)_{n} \right) = \sum_{n=1}^{N} \sum_{k=1}^{K} \one_k(C_n)\ln(p_{k}) +\one_k(C_n)\ln(\pr{X^{(k)}}(x_n; a_k)).
}
The cost of this completion is to introduce unknown class variables $(C_n)_n$ which makes the log-likelihood $\ell$ a random variable preventing its direct maximization. This is solved by the two steps of the EM algorithm. Given the previous parameter estimates $\boldsymbol{\theta}^{(t)} = \left(\mathbf{p}^{(t)}, \mathbf{a}^{(t)}\right)$, the E-step estimates the completed-data log-likelihood $Q$ by taking expectation of $\ell$ knowing $\boldsymbol{\theta}^{(t)}$
\eql{
Q(\boldsymbol{\theta},\boldsymbol{\theta}^{(t)}) = \EE_{C_n\vert (X_n)_n, \boldsymbol{\theta}}\left( \ell\left(\mathbf{p},\mathbf{a}; (x_n, C_n)_{n} \right) \vert (x_n)_n, \boldsymbol{\theta}^{(t)} \right).
}
In practice, the E-step amounts to computing for all $n\in \{1,\dots,N\}$, for all $k\in \{1,\dots,K\}$, the posterior class probabilities
\eql{
 \tau_{n,k}^{(t)} = \mathbb{P}_{C_n\vert X_n, \boldsymbol{\theta}}( k \vert x_n,\boldsymbol{\theta}^{(t)}) = \EE_{C_n\vert X_n, \boldsymbol{\theta}}(\one_k(C_n)\vert x_n, \boldsymbol{\theta}^{(t)}).
}
Then, the M-step updates the previous parameter estimates by maximizing $Q$ \ie
\eql{
\boldsymbol{\theta}^{(t+1)} = \uargmax{\boldsymbol{\theta}} Q(\boldsymbol{\theta},\boldsymbol{\theta}^{(t)}).
}
With an appropriate choice of the component distributions, for instance Gaussian or Exponential, closed-form maximum likelihood estimates of the component parameters $\mathbf{a}$ are available. In addition, the update rule for the mixing probabilities does not depend on the choice of the component distribution and it writes
\eql{\label{eq:mixing-prob-update-rule-standard}
p_{k}^{(t+1)} = \frac{1}{N}\sum_{n=1}^N \tau_{n,k}^{(t)}.
}

This procedure was introduced by Dempster \etal{} \cite{dempster1977maximum} which proved that the likelihood is non-decreasing at each iteration of the EM algorithm. There are no general guarantees that the sequence $\{ \boldsymbol{\theta}^{(t)}\}$ converges to a maximum likelihood estimator. Under some conditions verified by many models, the EM algorithm converges to a stationary value of the complete-data log-likelihood function while the convergence of the sequence $\{ \boldsymbol{\theta}^{(t)}\}$ to a point $\boldsymbol{\theta}^\ast$ requires stronger conditions \cite{wu1983convergence,boyles1983convergence}.

\subsection{Previous work and contributions}

As explained above, an important limitation of mixture models is that they assume independence between samples, and therefore ignore the underlying topology of the dataset. For instance, considering image segmentation, when the samples are pixels of an image, the location of a pixel and the classes assigned to its neighbors may provide information about the class assignment of that pixel. We consider the approach in which the mixing probabilities depend on the index $n$ of the sample (\ie{} $\mathbf{p}$ becomes $\mathbf{p}_n$), and therefore the mixture model writes
\eql{
\cpr{X_n}{\mathbf{P}_n,\mathbf{A}}(x\vert \mathbf{p}_n,\mathbf{a}) = \sum_{k=1}^K p_{n,k} \mathbb{P}_{X^{(k)}}(x;a_k),
\label{eq:pmm}}
where for all $n \in \{1, \dots, N\},\: \mathbf{p}_n \in \Delta^K$.

The model now is over-parametrized, but the topology of the data can be exploited to regularize the mixing probabilities. Specifically, our work is closely related to the approaches of Sun~\etal{}~\cite{sun2017location} and Nikou~\etal{}~\cite{nikou2010bayesian}. Therefore, in the following we summarize their work before introducing ours.

\paragraph{Preliminary definitions} First, we say that a random variable $R$ follows a Gamma distribution knowing the parameter $S$ when its density writes
\eql{
\label{eq:Gamma}
\cpr{R}{S}(r\vert s) = \frac{r^{s-1} \exp\left(-r\right)}{\Gamma(s)}.
}
We denote $R\sim \Gg(S)$. Then, we say that a random vector $\mathbf{R} \in \Delta^{K}$ follows a Dirichlet distribution knowing the parameters $\mathbf{S}$ when its density writes
\eql{
\label{eq:Dirichlet}
\cpr{\mathbf{R}}{\mathbf{S}}(\mathbf{r}\vert \mathbf{s}) = \frac{\Gamma \left( \sum_{k=1}^K s_k\right)}{ \prod_{k=1}^K \Gamma(s_k)} \prod_{k=1}^K r_{k}^{s_k-1}.
}
We denote $\mathbf{R}\sim \Dd(\mathbf{S})$. Then, a function $F: \RR^2 \longrightarrow \RR$ is a Gaussian process with mean $\mu: \RR^2 \longrightarrow \RR$ and covariance $\Sigma: \RR^2 \times \RR^2 \longrightarrow \RR$ when for any $N>0$ and any locations $(l_1,\dots,l_N)\in \RR^{2\times N}$,\: $(F(l_n))_n \sim \Nn\left(\bar \mu,\bar \Sigma \right)$ where $\bar \mu = (\mu(l_n))_n$ and $\bar \Sigma = (\Sigma(l_m,l_n))_{m,n}$. We denote $F\sim \Gg\Pp(\mu,\Sigma)$. The Gaussian process $F$ is a Gauss-Markov process when for all $(m,n) \in \{1,\dots,N\}^2$ such that $l_m \notin \Cc_n,\: \Sigma^{-1}(l_m,l_n) = 0$ where $\Cc_n$ is a neighborhood of $l_n$.  We denote $F\sim \Gg\Mm\Pp(\mu,\Sigma)$. The Gaussian process $F$ is stationary when $\mu$ is constant and $\Sigma(u,v) = \Sigma(u-v)$.  We denote $F\sim \Ss\Gg\Pp(\mu,\Sigma)$.

\input{figs/figure-1}

\paragraph{Location-dependent Dirichlet Process} In the work of Sun~\etal{}~\cite{sun2017location} the mixing probabilities are modeled as:
\eql{
P_{n,k} = \frac{Q_k \exp\left(F_k(l_n)\right)}{\sum_{i=1}^{K} Q_i \exp\left(F_i(l_n)\right)}
}
where
\eq{
\forall k \in \{1,\dots,K\}, \: Q_k \sim \Gg(b) \qandq F_k \sim \Ss\Gg\Pp(0,\Sigma)
}
with $P_{n,k} = P_{k}(l_n)$ and $b>0$. The associated graphical model is shown in Figure~\ref{fig:graph-lddp}.

In the work of Sun~\etal{}, model training is achieved using variational inference. However, for the sake of comparison we have derived the EM update rule for the mixing probabilities (Table~\ref{tab:update-comparison}, left). These equations highlight how information about neighboring pixels is combined to obtain mixing probabilities. This formulation, involving additional variables, is more complex and difficult to interpret than ours.

\paragraph{Spatially-varying mixtures} In the work of Nikou~\etal{}~\cite{nikou2010bayesian} the mixing probabilities are modeled as
\eql{
\mathbf{P}_n = \mathbf{P}(l_n) \sim \Dd(\mathbf{B}(l_n)) \qwhereq \forall k \in \{1,\dots,K\}, \: B_k \sim \Gg\Mm\Pp(0,\Sigma_k).
}
The associated graphical model is shown in Figure~\ref{fig:graph-sv-mix}. The EM update rule for the mixing probabilities consists in solving a third order polynomial equation (Table~\ref{tab:update-comparison}, middle), which can cause numerical instabilities.

\paragraph{Our model: Flexibly regularized mixture models (FlexMM)} We assume that the random vector $\mathbf{P}_n$ follows a Dirichlet distribution whose parameter $\mathbf{B}_n \in \mathbb{R}^K$ depends linearly on the classes $(C_n)_n$. In other words, the Dirichlet parameter pulls information from the classes of other samples to regularize $\mathbf{P}_n$. Our model for the mixing probabilities writes, for all $n \in \{1,\dots,N\}$,
\eql{\label{eq:our-model}
\mathbf{P}_n = \mathbf{P}(l_n) \sim \Dd(\mathbf{B}(l_n))
}
where
\eql{\label{eq:our-model-param}
\forall k \in \{1,\dots,K\}, \: B_k(l_n)= u_{n,k}((\one_k(C_n))_n) - \one_k(C_n) + 1
}
with $u_{n,k}: \RR^N \longrightarrow \RR$ is a linear function such that $u_{n,k}\left([0,+\infty[^N\right) \subset \RR_+$. The graphical model associated with our model is shown in Figure~\ref{fig:graph-ours-mixture}.

\revisedA{
Importantly, by defining parameters $\textbf{B}$ this way, the update rule depends linearly on the posterior class probabilities, in contrast to \cite{sun2017location,nikou2010bayesian}, and is entirely determined by the functions $u_{n,k}$ (Table~\ref{tab:update-comparison}, right; see Section \ref{sec:linear-update-mixing-prob} for the derivation). These functions are defined according to the data on which one wants to apply the model, and its underlying structure, as they determine how information can be combined and propagated between data points throughout the iterations of the EM algorithm. For instance, if the model is applied to the pixels of an image, one usually wants to combine information obtained from neighboring pixels. If the dataset gathers words in sentences, it can make sense to combine information from neighboring words but also from the importance or the position of the word in the sentence. Indeed, depending on the space on which is defined the dataset, functions $u_{n,k}$ can be chosen to favor the grouping of points in the sense of an adapted distance, entirely determined by the user. Some specific examples and an application to image segmentation are provided in Sections \ref{sec:linear-update-mixing-prob} and \ref{sec:numerical-exp}.
}

Therefore, our formulation has the advantage of guaranteeing linear updates while maintaining full flexibility in how class information is integrated across samples. Yet, this comes at the cost of introducing loops in the graphical model. In the following section, first, we justify that our loopy model is well-defined, providing a detailed description, and we state formally the theoretical results which comes with it. Then, we show how the flexible update rule allows one to combine multiple mixture models through their mixing probabilities.

\begin{table}[H]
    \centering
    \begin{tabular}{m{0.7cm}|m{4.2cm}|m{3cm}|m{3.8cm}}
       & \hspace{5mm} Sun~\etal{}~\cite{sun2017location}  & \centering Nikou~\etal{}~\cite{nikou2010bayesian} & {\hspace{15mm} Ours} \\
       \hline
     \vspace{-15mm}\rotatebox[origin=c]{90}{\parbox[c]{3.8cm}{\centering Mixing probabilities update}} & {\begin{align*} q_k^{(t+1)} &= b-1 + \sum_{n=1}^N \tau_{n,k}^{(t)}-p_{n,k}^{(t)} \\ f_{\cdot,k}^{(t+1)} &= \bar \Sigma (\tau_{\cdot,k}^{(t)}-p_{\cdot,k}^{(t)}) \\ p_{n,k}^{(t+1)} &= \frac{ q_k^{(t+1)} \exp\left(f_{n,k}^{(t+1)}\right) }{\sum_{i=1}^{K} q_i^{(t+1)} \exp\left(f_{n,i}^{(t+1)}\right)} \end{align*}}  & $p_{n,k}^{(t+1)}$ depends on the roots of 3$^{\text{rd}}$ order polynomials which depend on the posterior class probabilities $\tau_{n,k}^{(t)}$. & $$p_{n,k}^{(t+1)} = \frac{u_{n,k}(\tau_{\cdot,k}^{(t)})}{\sum_{k=1}^K u_{n,k}(\tau_{\cdot,k}^{(t)}) }$$
    \end{tabular}
    \caption{Comparison of the mixing probabilities update rule of our model to the ones of previous models. We use the notation $p_{n,k}=p_k(l_n)$. These update rules should be compared to the standard mixture update rule given in the previous subsection: $
p_{k}^{(t+1)} = \frac{1}{N}\sum_{n=1}^N \tau_{n,k}^{(t)}$.}
    \label{tab:update-comparison}
\end{table}

\section{Linear Update of the Mixing Probabilities}
\label{sec:linear-update-mixing-prob}

\subsection{Single mixture model}
\label{sec:single-mixture-model}

As explained in the previous section, the method we propose to regularize the over-parametrized mixing probabilities $\mathbf{p}_n$ introduces loops in our graphical model (Figure~\ref{fig:graph-ours-mixture}). Such loops could be problematic as they may be inconsistent or prevent one from defining a joint distribution. To circumvent these issues, we identify our directed graph with a factor graph.

A variant of the EM algorithm, called the factor graph EM algorithm \cite{eckford2000iterative}, was developed to obtain ML or MAP estimates for models associated with a graph containing cycles or loops. Factor graphs~\cite{frey2003extending,murphy2012machine} are graphical models introduced to explicitly represent arbitrary factorization of the joint distribution. A factor graph has two type of nodes. Variables, either known or hidden, are often represented by letters in circles and factors, which define relations between variables, are represented by black squares and are associated with factor functions given in the factorization. Eckford and Pasupathy~\cite{eckford2000iterative,eckford2004channel} state that this EM algorithm on factor graphs is able to break cycles and infer the parameters of the model. The key of this strategy is to extract from the initial factor graph two subgraphs. The first subgraph contains the hidden random variables of the model that are estimated during the E-step, and it is therefore called the E-step factor graph. The second subgraph contains the parameters of the model that are estimated during the M-step of the algorithm, and is called the M-step factor graph. Note that in that case, each step of the EM algorithm can also be seen as local message computations~\cite{dauwels2005expectation, dauwels2009expectation}, which are at the core of message passing algorithms. Eckford~\cite{eckford2004channel} shows that if both subgraphs are cycle-free, then the EM algorithm can be implemented exactly.
Eckford~\cite{eckford2004channel} and Dauwels~\etal~\cite{dauwels2005expectation} show that using a likelihood function computed from the model, one can define the function $Q$ adapted to the factor graph that will be computed and maximized iteratively during the EM algorithm.

Thanks to the Hammersley-Clifford theorem~\cite{koller2009probabilistic}, we know that the joint distribution of the hidden variables $C$, $\mathbf{B}$, the observations $X$ and the parameters $\mathbf{p},\mathbf{a}$ is proportional to the product of potential functions on the cliques of the associated graph. We choose the following factorization associated to our model
\eqALl{\label{eq:joint-distribution_alldata}
\pr{X,C,\mathbf{B},\mathbf{P},\mathbf{A}}&((x_n)_n,(C_n)_n,(\mathbf{B}_n)_n,(\mathbf{p}_n)_n,\mathbf{a}) \nonumber \\
& = \frac{1}{Z}f_{\mathbf{A}}(\mathbf{a}) \prod_{n=1}^N f_{X,C,\mathbf{A},\mathbf{P}}(x_n,C_n,\mathbf{a},\mathbf{p}_n) f_{\mathbf{P},\mathbf{B}}(\mathbf{p}_n, \mathbf{B}_n) f_{\mathbf{B},(C_n)_n}(\mathbf{B}_n,(C_n)_n),
}
where the potential functions are chosen to be equal to the following conditional probabilities,
\eqALl{&f_{X,C,\mathbf{A},\mathbf{P}}(x_n,C_n,\mathbf{a},\mathbf{p}_n) = \cpr{X,C}{\mathbf{A},\mathbf{P}}(x_n, C_n \vert \mathbf{a},\mathbf{p}_n) \label{eq:factor-xcap}\\
&f_{\mathbf{P},\mathbf{B}}(\mathbf{p}_n, \mathbf{B}_n) = \cpr{\mathbf{P}}{\mathbf{B}}(\mathbf{p}_n\vert \mathbf{B}_n)\\
&f_{\mathbf{B},(C_n)_n}(\mathbf{B}_n,(C_n)_n)=\cpr{\mathbf{B}}{(C_n)_n}(\mathbf{B}_n\vert (C_n)_n)\\
&f_{\mathbf{A}}(\mathbf{a}) =  \pr{\mathbf{A}}(\mathbf{a}),\label{eq:factor-a}
}
and where $Z$ is normalization constant. Figure \ref{fig:factographall} presents the factor graph associated with this factorization and our initial model, using the same notations as the seminal papers on factor graphs~\cite{kschischang2001factor,frey2003extending}. The E-step of the EM algorithm computes an expectation given the parameters estimated during the previous iteration, which correspond to an implicit inference of the hidden variables $(C_n)_n$ and $(\mathbf{B}_n)_n$. The M-step corresponds to the maximization of this expectation to obtain new estimates for $\mathbf{p}$ and $\mathbf{a}$. Given the factorization and the computations done during the E-step and the M-step, both subgraphs associated with each step of the EM algorithm are presented in Figure \ref{fig:factorgraphs}. Note that there is no loop in either subgraph. Thus, the proofs in \cite{eckford2000iterative,dauwels2005expectation} still hold and the EM algorithm is guaranteed to iteratively increase the incomplete log-posterior.

\input{figs/figure-2}

The estimation of the component parameter $\mathbf{a}$ is independent from the estimation of the mixing probabilities $\mathbf{p}_{n}$. Therefore, any prior $\pr{\mathbf{A}}$ can be used for the component parameters $\mathbf{a}$. In the EM approach, we consider the joint probability of $x_n$ and $C_n$. Hence, the log-posterior distribution of $(\mathbf{p},\mathbf{a})$ writes
\eqALl{
\ell(\mathbf{p},\mathbf{a} ;& (x_n, C_n,\mathbf{B}_n)_{n}) = \ln(\cpr{\mathbf{P},\mathbf{A}}{X,C,\mathbf{B}}(\mathbf{p},\mathbf{a}\vert (x_n)_n,(C_n)_n,(\mathbf{B}_n)_n) ) \nonumber\\
=& \ln(\pr{X,C,\mathbf{B},\mathbf{P},\mathbf{A}}((x_n)_n,(C_n)_n,(\mathbf{B}_n)_n,(\mathbf{p}_n)_n,\mathbf{a})) - \ln(\pr{X,C,\mathbf{B}}((x_n)_n,(C_n)_n,(\mathbf{B}_n)_n)    \nonumber\\
= & \: \sum_{n=1}^N \ln(\cpr{X,C}{\mathbf{P},\mathbf{A}}(x_n,C_n \vert \mathbf{p}_n,\mathbf{a})) +  \ln( \cpr{\mathbf{P}}{\mathbf{B}}(\mathbf{p}_{n}\vert\mathbf{B}_n)) + \ln(\cpr{\mathbf{B}_n}{(C_n)_n}(\mathbf{B}_n\vert (C_n)_n ))\\
&+ \ln( \pr{\mathbf{A}}(\mathbf{a}) )- \ln(\pr{X,C,\mathbf{B}}((x_n)_n,(C_n)_n,(\mathbf{B}_n)_n) - \ln(Z) . \nonumber
}
We recall that we only observe realizations of $X$. Therefore in the EM approach, the unobserved pixel classes $(C_n)_n$ and prior parameters $(B_n)_n$ will be estimated by taking the expectation during the E-step. In this mixture model, the expression of $\cpr{X,C}{\mathbf{P},\mathbf{A}}$ used in Equation~\eqref{eq:log-lkl} still holds. Then,
\eql{
\label{eq:JointXC}
\cpr{X,C}{\mathbf{P},\mathbf{A}}(x_n, C_n \vert \mathbf{p}_n; \mathbf{a}) =  \prod_{k=1}^{K} \left(p_{n,k} \pr{X^{(k)}}(x_n; a_k) \right)^{\one_k(C_n)}\\
}
In addition, we assume that $\mathbf{B}_n$ is entirely characterized by the classes $(C_n)_n$ \ie{}  it has the following Dirac density
\eql{\label{eq:def-dir-param}
\cpr{\mathbf{B}_n}{(C_n)_n}(\mathbf{B}\vert (C_n)_n ) = \delta_{\mathbf{v}_n\left(\one_1(C_\cdot),\dots,\one_K(C_\cdot)\right)}(\mathbf{B}),
}
where $\delta_x(y)$ is the Dirac delta distribution and where $\mathbf{v}_n: \RR^{N\times K} \mapsto \RR^K$ is a linear function. This amounts to consider that $\mathbf{B}_n$ is equal in distribution to $\mathbf{v}_n\left(\one_1(C_\cdot),\dots,\one_K(C_\cdot)\right)$. In particular, when $\mathbf{v}_n\left(\one_1(C_\cdot),\dots,\one_K(C_\cdot)\right) = (u_{n,1}(\one_1(C_\cdot)) - \one_1(C_n) + 1,\dots, u_{n,K}(\one_K(C_\cdot)) - \one_K(C_n) + 1 )$, this comes down to assume that the distribution of the probabilistic maps $\mathbf{P}_n$ is given by Equation~\eqref{eq:our-model} with parameters given by Equation~\eqref{eq:our-model-param}.
Finally the log-posterior can be simplified as
\eqALl{\label{eq:lkl}
\ell(\mathbf{p},\mathbf{a};&(x_n, C_n,\mathbf{B}_n)_{n}) = \sum_{n=1}^{N}  \sum_{k=1}^{K} \one_k(C_n) \Big[\ln\left( p_{n,k}  \right) + \ln\left( \pr{X^{(k)}}(x_n; a_k) \right)\Big] + \ln( \pr{\mathbf{A}}(\mathbf{a}) )\nonumber\\
& + \ln\left(  \cpr{\mathbf{P}}{\mathbf{B}}\left(\mathbf{p}_{n}\Big\vert\mathbf{v}_n\left(\one_1(C_\cdot),\dots,\one_K(C_\cdot)\right)\right)\right) + W((x_n, C_n,\mathbf{B}_n)_{n}),
}
where $W$ is the function that gathers all the terms of $\ell$ that do not depend on $\mathbf{p}$ or $\mathbf{a}$.
From there assuming that $\cpr{\mathbf{P}}{\mathbf{B}}$ is a Dirichlet distribution as given by Equation~\eqref{eq:Dirichlet}, it is possible to derive a custom update rule for the mixing probabilities which is applicable to any mixture model as stated in the following proposition.

\begin{prop}
\label{prop:gen-update-rule}
For all $(n,k) \in \Omega_{N,K}=\{1,\dots,N\}\times \{1,\dots,K\},$ let $u_{n,k}: \RR^N \longrightarrow \RR$ be any linear function such that $u_{n,k}\left([0,+\infty[^N\right) \subset \RR_+$. If $\mathbf{v}_n\left(\one_1(C_\cdot),\dots,\one_K(C_\cdot)\right) = (u_{n,1}(\one_1(C_\cdot)) - \one_1(C_n) + 1,\dots, u_{n,K}(\one_K(C_\cdot)) - \one_K(C_n) + 1 ) $, then, the mixing probability updates are
\eql{\label{eq:gen-update-rule}
\forall (n,k) \in \Omega_{N,K},  \quad p_{n,k}^{(t+1)} = \frac{u_{n,k}(\tau_{\cdot,k}^{(t)})}{\sum_{k=1}^K u_{n,k}(\tau_{\cdot,k}^{(t)}) }.
}
where $\tau_{n,k}^{(t)}=\cpr{C_n}{X_n,\boldsymbol{\Theta}}(k|x_n,\boldsymbol{\theta}^{(t)})$ is the $k^{\text{th}}$ component posterior probability of sample $x_n$ computed at the previous E-step and $\boldsymbol{\theta}^{(t)}$ is the previous parameter estimate.
\end{prop}
\begin{proof}
The proof has two steps: (i) take the conditional expectation of the log-posterior~\eqref{eq:lkl} knowing the data and the parameters estimated at the last M-step and use the equality $\EE(({B}_{n,k}-1 + \one_k(C_n))\vert (x_n)_n,\boldsymbol{\theta}^{(t)}) = u_{n,k}(\tau_{\cdot,k}^{(t)})$; (ii) write the Karush–Kuhn–Tucker condition~\cite{boyd2004convex} for $p_{n,k}$. See supplementary section~\ref{asec:proof} for details.
\end{proof}
Defining the distribution of the Dirichlet parameters $\textbf{B}$ as in \eqref{eq:def-dir-param} enables to adapt the way spatial information is propagated, given a set of linear functions $(u_{n,k})_{n,k}$. It also leads to linear update equations~\eqref{eq:gen-update-rule} which was our initial goal. These functions can be defined according to the application of the model (such as image segmentation, text categorization, scene classification) and may also depend on the position of the sample they are applied to.
In particular, when $u_{n,k}(\tau_{\cdot,k}) = \sum_m \tau_{m,k}/N$, the update corresponds to the standard mixture model. When $u_{n,k}(\tau_{\cdot,k}) = \tau_{n,k}$, the mixing probabilities will be equal to the component posterior probability of $x_n$. Finally, when $u_{n,k}(\tau_{\cdot,k}) = {G \ast \tau_{.,k}}_{\vert n}$ where $G$ is any averaging kernel and $\ast$ is a convolution operator, both adapted to the topology of indexes $n$, the update corresponds to a local average of the posterior as has been used recently for spatial smoothing~\cite{vacher2018probabilistic,he2010laplacian}.
Despite the reformulation of our model using factor graphs, we are still able to compute the quantities needed to perform an EM algorithm and we can prove the following result.
\begin{prop}
\label{prop:EM-increases-logpost}
The EM algorithm applied to this model is an ascent algorithm, meaning that at each iteration $t+1$, the incomplete log-posterior increases as the parameters are updated:
\eq{\ln \cpr{\mathbf{P},\mathbf{A}}{X}(\mathbf{p}^{(t+1)},\mathbf{a}^{(t+1)}\vert (x_n)_n)  \geq \ln \cpr{\mathbf{P},\mathbf{A}}{X}(\mathbf{p}^{(t)},\mathbf{a}^{(t)}\vert (x_n)_n).
}
\end{prop}
The proof of this proposition adapts the proof of the monotone behavior of log-likelihood of the usual EM algorithm~\cite{dempster1977maximum} to our model, it can be found in Supplementary Section~\ref{asec:increase}. Similarly, the convergence properties of the EM algorithm given in~\cite{wu1983convergence} remain valid in our framework, thus the log-posterior function converges to a stationary point.

\subsection{Combining mixture models}
\label{sec:multi-mixture-models}

Real datasets are often composed of multiple feature vectors that have different dimensionality and different number of samples. In addition, multiple feature vectors can be organized according to a natural topology, \eg{} the activations at different layers of DNNs correspond to different feature vectors and are organized hierarchically. In such settings, a simple approach could be to train several mixture models independently. However, doing so would ignore the topological organization of the feature vectors, and furthermore the interchangeability of mixture components will prevent to gather the results without relabelling. Relabelling is a costly (non-polynomial) counting problem.

We propose here to alleviate these issues by sharing class information across different mixture models, using a strategy similar to the one introduced in the previous Section. The idea is that instead of training multiple mixture models independently, these can be trained in parallel and regularize each other through the mixing probabilities (see Figure~\ref{fig:modified-mixture-multi}). As we show below, the update of the mixing probabilities is linear and can be fully specified through the $\mathbf{v}_n^{(h)}$ function, allowing one to decide how to combine the class information from different mixture models.

We assume that each data sample is associated with a collection of features vectors $(x_n^{(h)})_{h \in \{1, \dots, H \}}$ indexed by the index $h$. In the case of images, $h$ denotes the DNN layers which contain the activation vector to an image at each pixel location $l_n$. Again, the Dirichlet parameters $\mathbf{B}^{(h)}_n$ follow a Dirac distribution
\eql{\label{eq:def-dir-param-multi}
\cpr{\mathbf{B}^{(h)}_n}{(C^{(h)}_n)_{n,h}}(\mathbf{B}\vert (C^{(h)}_n)_{n,h} ) = \delta_{\mathbf{v}_n^{(h)}\left(\one_1({C_\cdot^{(1)}}),\dots,\one_1({C_\cdot^{(H)}}),\dots,\one_K({C_\cdot^{(1)}}),\dots, \one_K({C_\cdot^{(H)}})\right)}(\mathbf{B})
}
where $\mathbf{v}_n^{(h)}: \RR^{N \times H \times K} \mapsto \RR^K$ is a linear function. As such, this function will set the Dirichlet parameters so that it will regularize the mixing probability $\mathbf{p}^{(h)}_n$ with the class information of the neighboring samples and layers.

\input{figs/figure-3.tex}

As in the previous section, we estimate the parameters of our model using MAP inference
\eql{\label{eq:map-layers}
(\hat{\mathbf{a}}, \hat{\mathbf{p}}) = \uargmax{(\mathbf{a}, \mathbf{p})}  \; \ell\left(\mathbf{p},\mathbf{a}; (x^{(h)}_n, C^{(h)}_n,\mathbf{B}^{(h)}_n)_{n,h} \right)
}
where
\eql{\label{eq:log-post-layers}
\ell(\mathbf{p},\mathbf{a}; (x^{(h)}_n, C^{(h)}_n,\mathbf{B}^{(h)}_n)_{n,h} )=\ln  \left( \mathbb{P}_{\mathbf{A}, \mathbf{P} \vert X,C, \mathbf{B}}(\mathbf{a}, \mathbf{p} \vert (x^{(h)}_n)_{n,h}, (C^{(h)}_n)_{n,h}, (\mathbf{B}^{(h)}_n)_{n,h}) \right).
}
Assuming additionally that the feature and class variables $(x^{(h)}_n,C^{(h)}_n)_h$ are independent knowing $(\mathbf{p}^{(h)}_n,\mathbf{a}^{(h)})_h$ and that the mixing probabilities $(\mathbf{p}^{(h)}_n)_h$ are independent knowing $(\mathbf{B}^{(h)}_n)_h$, the posterior in Equation~\eqref{eq:log-post-layers} can be factorized in a very similar way as in the previous section. Indeed, the log-posterior writes
\eqALl{\label{eq:lkl-layers}
\ell(\mathbf{p},\mathbf{a};& (x^{(h)}_n, C^{(h)}_n,\mathbf{B}^{(h)}_n)_{n,h} )  = \sum_{n=1}^{N} \sum_{h=1}^H \sum_{k=1}^{K} \one_k(C^{(h)}_n) \Big[\ln\left( p^{(h)}_{n,k}  \right) + \ln\left( \pr{X^{(h,k)}}(x^{(h)}_n; a^{(h)}_k) \right)\Big] \nonumber\\
& + \ln\left(  \cpr{\mathbf{P}}{\mathbf{B}}\left(\mathbf{p}^{(h)}_{n}\Big\vert \mathbf{v}_n^{(h)}\left(\one_1({C_\cdot^{(1)}}),\dots,\one_1({C_\cdot^{(H)}}),\dots,\one_K({C_\cdot^{(1)}}),\dots, \one_K({C_\cdot^{(H)}})\right) \right)\right) \nonumber\\
& + \sum_{h=1}^H \pr{\mathbf{A}}(\mathbf{a}^{(h)})   + W((x^h_n,C^h_n,\mathbf{B}^h_n)_{n,h}),
}
where $W$ is the function gathering the quantities in $\ell$ that do not depend on $\mathbf{p}$ or $\mathbf{a}$.
As in Proposition~\ref{prop:gen-update-rule}, we can derive linear update rules for the probability maps in the EM algorithm as detailed in the following proposition.
\begin{prop}
\label{prop:gen-update-rule-ext}
For all $(n,k,h) \in \Omega_{N,K,H}=\{1,\dots,N\}\times \{1,\dots,K\}\times \{1,\dots,H\},$ let $u_{n,k}^{(h)}: \RR^N \longrightarrow \RR$ be any linear function such that $u_{n,k}^{(h)}\left([0,+\infty[^{N H}\right) \subset \RR_+$. If
\eqAL{\mathbf{v}_n^{(h)}(&\one_1(C_\cdot^{(1)}),\dots,\one_1(C_\cdot^{(H)}),\dots,\one_K(C_\cdot^{(1)}),\dots, \one_K(C_\cdot^{(H)}))\\
& =  (u_{n,1}^{(h)}(\one_1(C_\cdot^{(1)}),\dots,\one_1(C_\cdot^{(H)})) - \one_1(C_\cdot^{(h)}) + 1,\dots,\\
& \quad\quad u_{n,K}^{(h)}(\one_K(C_\cdot^{(1)}),\dots,\one_K(C_\cdot^{(H)})) - \one_K(C_\cdot^{(h)}) + 1 )}
then, the mixing probability updates of layer $h$ are
\eql{\label{eq:gen-update-rule-ext}
\forall (n,k,h) \in \Omega_{N,K,H},  \quad  p_{n,k}^{(t+1,h)} = \frac{u_{n,k}^{(h)}(\tau_{\cdot,k}^{(t,1)},\dots,\tau_{\cdot,k}^{(t,H)})}{\sum_{k=1}^K u_{n,k}^{(h)}(\tau_{\cdot,k}^{(t,1)},\dots,\tau_{\cdot,k}^{(t,H)}) }
}
where $\tau_{n,k}^{(t,h)}=\cpr{C_n^{(h)}}{X_n^{(h)},\boldsymbol{\Theta}}(k|x_n^{(h)},\boldsymbol{\theta}^{(t,h)})$ is the $k^{\text{th}}$ component posterior probability of sample $x_n^{(h)}$ at the previous E-step and $\boldsymbol{\theta}^{(t,h)}$ is the previous parameter estimate.
\end{prop}
The proof of this proposition is similar to the proof of Proposition \ref{prop:gen-update-rule} and can be found in supplementary section \ref{asec:proof}. Moreover, it is also possible to derive two sub-factor graphs from this model, each one associated with a step of the EM algorithm with no loop, so each iteration of the EM algorithm does increase the incomplete log-posterior. The pseudo code implementing our model for combining multiple mixture models is given in
Algorithm~\ref{alg:EMlayers}.
\setlength{\textfloatsep}{2mm}%
\begin{algorithm}[tb]
\textbf{Input:} Data feature vectors $(x_n^{(h)})_{n,h}$, numbers of iterations $n_{\text{iter}}$, of components $K$ and of layers $H$, linear functions $(u_{n,k}^{(h)})_{n,h}$.\\
\textbf{Output:} Mixing probability maps $\hat{\mathbf{p}}^{(h)}$ and mixture parameters $\hat{\mathbf{a}}^{(h)}$.
\begin{enumerate}
\item Initialize mixture parameters of layer 1 with K-means algorithm.
\item Initialize mixing probabilities of other layers with the posterior probabilities of layer 1.
\item Run M-step for all layer $h \geq 1$.
\item For $t \leq n_{iter}$,
\begin{itemize}
\item E-step:\\
For $h \leq H$, compute $\tau_{n,k}^{(t,h)}$.
\item M-step:\\
For $h \leq H$, compute mixing probability maps $p_{n,k}^{(t+1,h)}$ using Equation \eqref{eq:gen-update-rule-ext} and model parameters $\mathbf{a}^{(t+1,h)}$.
\end{itemize}
\end{enumerate}
\caption{MAP inference on a probabilistic model combining mixture models.}
\label{alg:EMlayers}
\end{algorithm}

\section{Numerical Experiments}
\label{sec:numerical-exp}
\subsection{Application to Synthetic Image Segmentation}

\subsubsection{Single Mixture}

\paragraph{Implementation details} We first validated our algorithms on synthetic data generated from a mixture model with $K=3$ components and Gaussian samples per component. We arranged observations on a 2D grid of size $N=256\times256$ with feature dimension $D=3$, similar to an RGB image with 256$\times$256 pixels. To generate synthetic data compatible with the model of Figure~\ref{fig:graph-ours-mixture}, we started from deterministic component--assignment maps (\ie{} $\mathbf{P}_n$ is a one-hot vector), obtained by manually partitioning the image in 3 spatially compact regions.  Then, we sampled the component assignments $\mathbf{C}$ and observations ${X}$ from \eqref{eq:JointXC}.

\begin{figure}
    \centering
    \begin{tikzpicture}
        \draw[step=1.0,white,thin] (-7.5,-5) grid (7.25,6.0);
        \node at (-2.75,5.5){\textbf{c)}};
       	\node at (2.5,0.1){\includegraphics[width=0.6\linewidth]{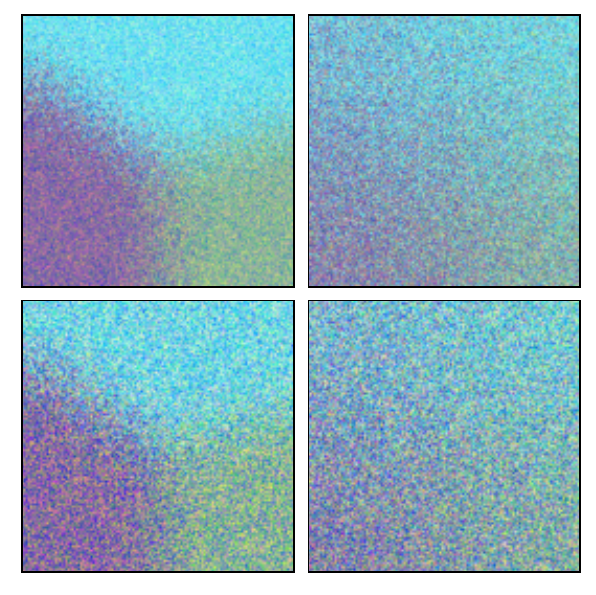}};
       	\draw[thick,-T] (-2.25,4.75)--(7.25,4.75);
       	\draw[thick,-T] (-2.25,4.75)--(-2.25,-4.6);
       	\node[text width=5.5cm] at (2.75,5.0){Prior Location Uncertainty};
       	\node[text width=6.5cm, rotate=90] at (-2.55,0.5){Component Overlap Uncertainty};

       	\node at (-7.25,5.5){\textbf{a)}};
       	\node[text width=3cm, align=center] at (-5.3,5){Prior Location Uncertainty};
       	\begin{axis}[every axis plot post/.append style={
          mark=none,domain=0:7,samples=50,smooth, thick},
          axis line style = {->,thick},
          height=5cm,
          width=5cm,
          axis x line*=bottom,
          axis y line*=left,
          enlargelimits=upper,
          at={(-7cm,1.0cm)},
          xtick=\empty,
          xticklabels={,,},
          xlabel={Location ($n$)},
          ytick={1},
          yticklabels={1},
          ylabel={Probability ($\mathbf{P}_n$)},
          ylabel style={at={(-0.075,0.5)}}]
          \addplot[red] {sigmoid(3.5,0.25)};
          \addplot[red] {sigmoid(3.5,-0.25)};
          \addplot[blue] {0.4+0.2*sigmoid(3.5,0.25)};
          \addplot[blue] {0.4+0.2*sigmoid(3.5,-0.25)};
        \end{axis}

       	\node at (-7.25,0.0){\textbf{b)}};
       	\node[text width=3.5cm, align=center] at (-5.3,-0.5){Component Overlap Uncertainty};
       	\begin{axis}[every axis plot post/.append style={
          mark=none,domain=0:7,samples=50,smooth, thick},
          axis line style = {->,thick},
          height=5cm,
          width=5cm,
          axis x line*=bottom,
          axis y line*=left,
          enlargelimits=upper,
          at={(-7cm,-4.5cm)},
          xtick=\empty,
          xticklabels={,,},
          xlabel={Feature ($X$)},
          ytick=\empty,
          yticklabels={,,},
          ylabel={Probability ($\PP_{X^{(k)}})$},
          ylabel style={at={(-0.075,0.5)}},
          legend style={at={(0.7,0.95)},anchor=north}]
          \addplot[red] {gauss(1.0,0.25)};
          \addplot[red] {gauss(2.0,0.25)};
          \addplot[blue] {gauss(4.5,0.5)};
          \addplot[blue] {gauss(5.5,0.5)};
          \legend{low,,high}
        \end{axis}
        \begin{scope}[yshift=-12cm]
            \draw[step=1.0,white,thin] (-7.5,-6) grid (7.25,6.0);

            \node at (-7.25,6.5){\textbf{d)}};
            \node at (0,6.15){Prior Location Uncertainty};
            \node[rotate=90] at (-6.25,0){Component Overlap Uncertainty};
            \draw[thick,-T] (-6.0,5.9)--(6.0,5.9);
            \draw[thick,-T] (-6.0,5.9)--(-6.0,-6.0);
            \node[rotate=-90] at (5.9,4.75){GT};
            \node[rotate=-90] at (5.9,2.9){Ours};
            \node[rotate=-90] at (5.9,1.15){GMM};

            \node[rotate=-90] at (5.9,-1.15){GT};
            \node[rotate=-90] at (5.9,-2.9){Ours};
            \node[rotate=-90] at (5.9,-4.75){GMM};

            \node at (-0.15,2.9){
                 \includegraphics[width=0.38\linewidth]{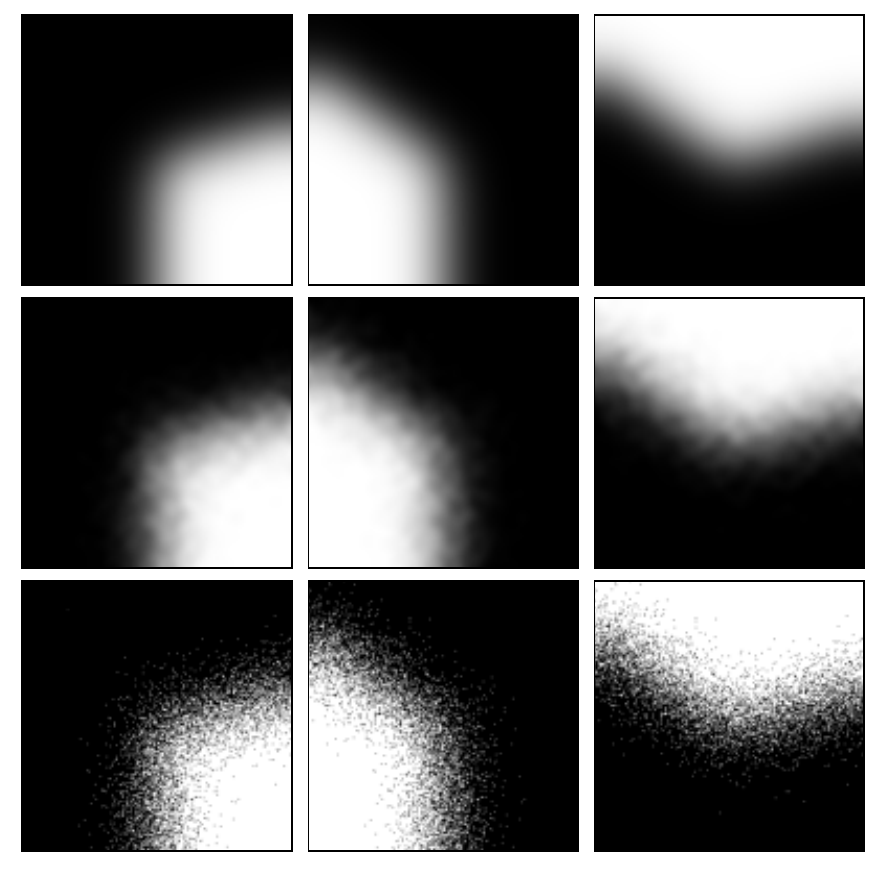}
                \includegraphics[width=0.38\linewidth]{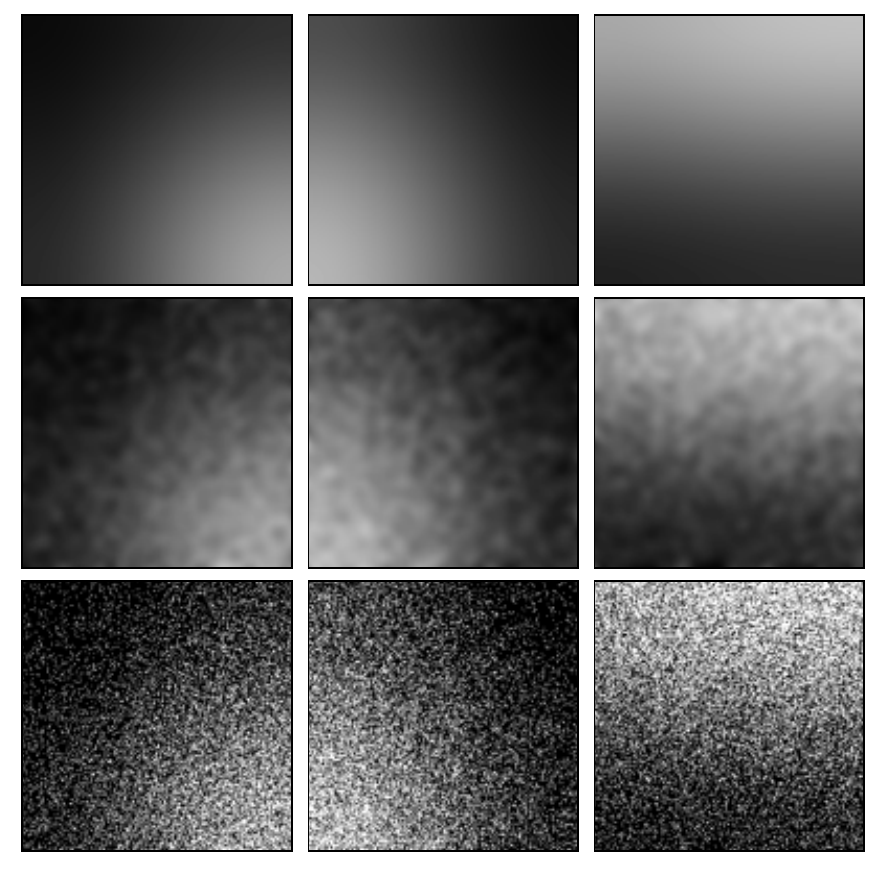}
                };
            \node at (-0.15,-2.9){
                \includegraphics[width=0.38\linewidth]{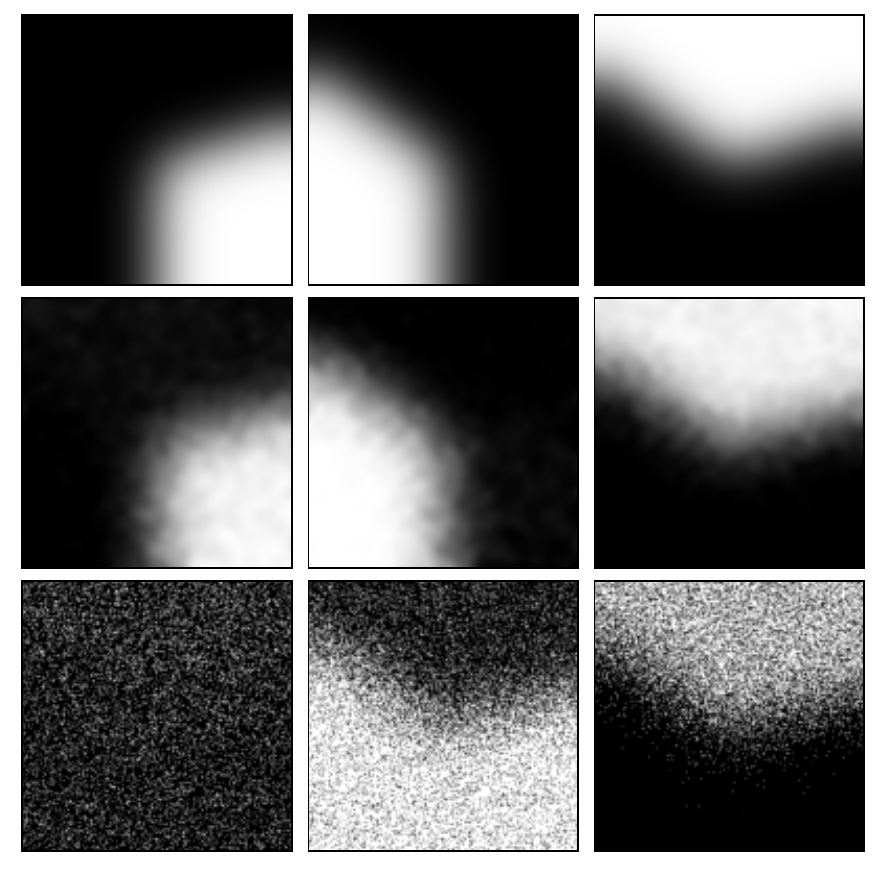}
                \includegraphics[width=0.38\linewidth]{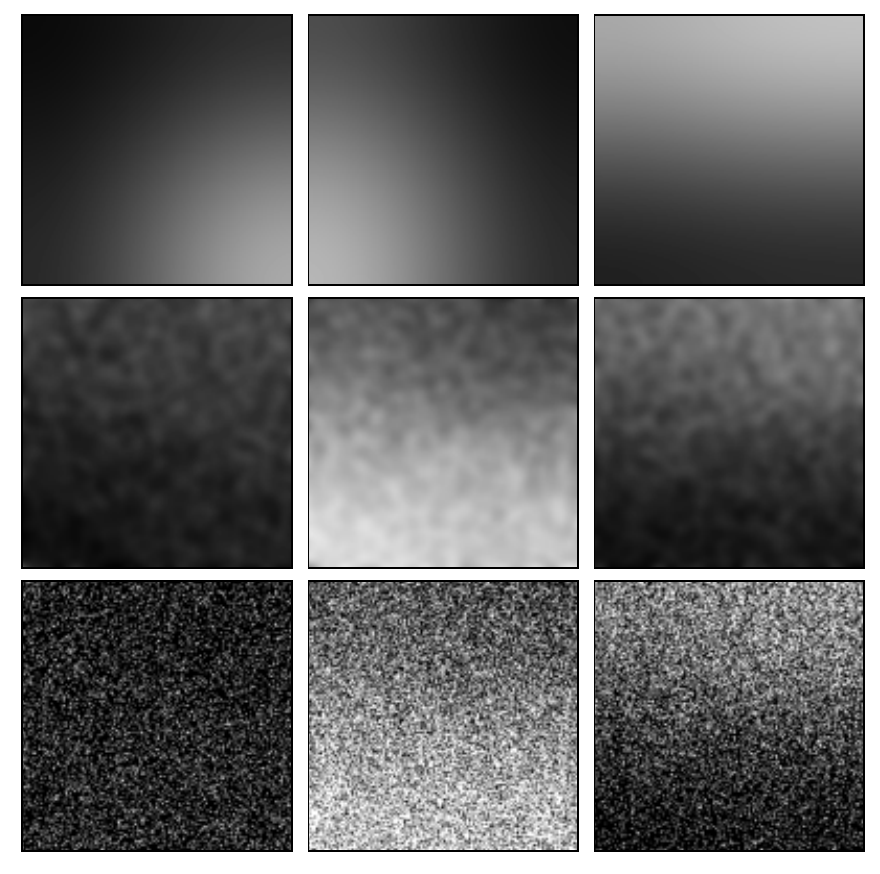}
                };
        \end{scope}
    \end{tikzpicture}
    \caption{See caption on the next page.}
    \label{fig:artif-data}
\end{figure}

One important test of our model is how it handles uncertainty about component assignment. Therefore, we performed two manipulations of the model parameters, to control uncertainty (Figure~\ref{fig:artif-data}). First, we controlled prior location uncertainty by systematically smoothing the maps $\mathbf{P}$ using a Gaussian function with increasing width (see Figure~\ref{fig:artif-data}a and Figure~\ref{fig:results-fit-multi-mixture}a). Second, we increased the overlap between the Gaussian distributions of the different classes, by increasing the observation variance (Figure~\ref{fig:artif-data}b). Figure~\ref{fig:artif-data}c illustrates one sample image for each of the 9 uncertainty combinations we explored, with location uncertainty increasing from left to right, and component overlap increasing from top to bottom.

\revisedA{
We applied our algorithm using 2-D Gaussian functions for $u_{n,k}$ (See Proposition \ref{prop:gen-update-rule}) to obtain a local spatial smoothing of the posterior. Hence, the update rule is
}
\eql{\label{eq:loc}
p_{n,k}^{(t+1)}  = G \ast \tau_{k}^{(t)}(l_n)
}
where $G$ is a Gaussian kernel with width $\sigma=5.25$, $\tau_{k}^{(t)}: l_n \mapsto \tau_{n,k}^{(t)}$ is the posterior maps and $\ast$ denotes the discrete convolution. For comparison, we also used a standard Gaussian Mixture Model with three components.

\paragraph{Results} We analyzed the component probability maps learned by our model (Figure~\ref{fig:artif-data}d, labeled ``Ours''; each map corresponds to one component, and lighter gray scale values correspond to higher probability of that component), and compared them to the ground truth maps (``GT'') as well as to a standard Gaussian Mixture model (``GMM''). When uncertainty is minimal, our algorithm recovers the GT maps accurately, and it improves slightly over the GMM, by removing isolated outliers thanks to the spatial smoothing (Figure~\ref{fig:artif-data}d, top--left block). Increasing uncertainty either by location prior or component overlap, leads to more dramatic failures of the GMM, whereas our model captures both the shape of the GT maps, as well as the increased uncertainty (abundance of intermediate gray levels, corresponding to probabilities close to 0.5, in the top row and left column of Figure~\ref{fig:artif-data}d). Importantly, when prior location uncertainty is low, our model is more robust to component overlap uncertainty than the GMM (left column and bottom blocks), whereas it displays similar failures as the GMM (except for the spatial smoothing) when both component overlap and prior location uncertainties become too large (Figure~\ref{fig:artif-data}d, bottom--right block).

In summary, together our simulations demonstrate that our algorithm learns probabilistic component-assignment maps, correctly capturing ground--truth labels and their uncertainty, and performing spatial smoothing. We note that the spatial smoothing achieved by our algorithm is qualitatively similar to that obtained by other approaches~~\cite{nikou2010bayesian,sun2017location}. However, our formulation also allows for straightforward and efficient integration of information consistent with different data topologies, well beyond spatial smoothing, which we illustrate next.

\subsubsection{Hierarchical Mixtures}

\paragraph{Implementation details} To illustrate the flexibility of our approach, we applied it to data generated from a 3--layer model as in Figure~\ref{fig:modified-mixture-multi}. This can be interpreted as a simple hierarchical model of images, with features corresponding to different layers of a DNN. For these simulations, we used three feature channels at each layer, and displayed the data as RGB images (Figure~\ref{fig:results-fit-multi-mixture}a right). The model at each layer is as in the previous section, with identical prior maps at all layers, except that the location uncertainty is largest in the first layer and smallest in the third layer (illustrated by the Ground Truth Maps in Figure~\ref{fig:results-fit-multi-mixture}a center). To perform the prior maps inference, we applied the model of Figure~\ref{fig:modified-mixture-multi} (\ie{} where component--assignments at one layer influence the prior of the other neighboring layers). \revisedA{In other words, we applied our algorithm defining $u_{n,k}^{(h)}$ as a combination of three Gaussian smoothing functions, each one operating a local spatial average of the posterior maps of the current, previous and next layers. This leads to the following update rule}
\eql{\label{eq:prior-update-a}
p_{n,k}^{(h,t+1)} = \frac{ {s_n^{(h,t)}}^2 {s_n^{(h+1,t)}}^2 m_{n,k}^{(h-1,t)} + {s_n^{(h-1,t)}}^2 {s_n^{(h+1,t)}}^2 m_{n,k}^{(h,t)} + {s_n^{(h-1,t)}}^2 {s_n^{(h,t)}}^2 m_{n,k}^{(h+1,t)} }{ {s_n^{(h,t)}}^2 {s_n^{(h+1,t)}}^2 + {s_n^{(h-1,t)}}^2 {s_n^{(h+1,t)}}^2 + {s_n^{(h-1,t)}}^2 {s_n^{(h,t)}}^2 },
}
where

\noindent
\begin{minipage}{0.36\linewidth}
\eq{\label{eq:loch}
m_{n,k}^{(h,t)}  = G^{(h)} \ast \tau_{k}^{(h,t)}(l_n),
}
\end{minipage}
\hspace{1mm}
\begin{minipage}{0.62\linewidth}
\eql{\label{eq:scale}
{s_n^{(h,t)}}^2 = \frac{\sum_{k=1}^K G^{(h)} \ast {\tau_{k}^{(h,t)}}^2(l_n) -{m_{n,k}^{(h,t)}}^2}{K(1-G^{(h)}\ast G^{(h)}(0))},
}
\end{minipage}
are respectively the local mean and variance of the posterior maps at layer $h$; $\tau_{k}^{(h,t)}: l_n \mapsto \tau_{n,k}^{(h,t)}$ is the posterior map at layer $h$; and $G^{(h)}$ is a 2-D Gaussian kernel with width $\sigma^{(h)}=5.25$.

\paragraph{Results} Figure~\ref{fig:results-fit-multi-mixture}b illustrates that our algorithm learned maps consistent with the ground truth at all layers, and also qualitatively reflected the decreasing uncertainty across layers. Importantly, because the integration across layers of assignment information is weighted by the relative uncertainties, the maps in the first layer are less uncertain than the corresponding ground truth (average difference between the entropy of the Fit and Ground Truth maps = -0.14), and, correspondingly, the maps in the second and third layer are more uncertain than the ground truth (entropy difference = 0.13 for layer 2, and 0.23 for layer 3).

\addtocounter{figure}{-1}
\begin{figure}[t!]
  \caption{Validation on synthetic color images with uncertainty manipulation. \textbf{a)} We control location uncertainty by setting the probability of each component to values close to 1 (red, low uncertainty) or to 0.5 (blue, high uncertainty), at any given location in the image (illustrated for a 1-dimensional image in the cartoon). \textbf{b)} We control component uncertainty by decreasing (red, low uncertainty) or increasing (blue, high uncertainty) the overlap of the Gaussian distributions of the two features within each component (illustrated for a 1-dimensional feature). \textbf{c)} Example synthetic images with three feature channels (corresponding to RGB values) and three components, generated from the model with increasing location uncertainty from left to right, and  and increasing component overlap from top to bottom. \textbf{d)} Each block of 3 by 3 maps contains the ground truth component probability maps (top row, labeled GT), and the maps recovered by our algorithm (middle row, labeled Ours) and by a Gaussian Mixture Model (bottom row, labeled GMM). Each map corresponds to a different component, and lighter pixels correspond to higher probability that the pixel is assigned to that component. Different 3 by 3 blocks correspond to different levels of location uncertainty and component overlap uncertainty, as described in \textbf{a)} and \textbf{b)}. Figure is best seen in color.}
\end{figure}

As a control, we also applied our single--mixture model independently to each layer (Figure ~\ref{fig:results-fit-multi-mixture}c). As expected, the algorithm recovered the maps correctly in the less uncertain layer, but was not able to recover them in more the uncertain layers (due to the high component uncertainty regime).

In summary, these results on synthetic data illustrate that our approach can integrate information about mixture components from multiple sources in a principled way, and easily adapt to different topologies of the data (in our examples, spatial arrangement and hierarchical organization).

\begin{figure}
    \centering
    \begin{tikzpicture}
        \draw[step=1.0,white,thin] (-7.5,14.5) grid (7.5,2.8);

        \node at (-7.25,14.0){\textbf{a)}};
        \node at (-5.2,8){\textbf{b)}};
        \node at (0.75,8){\textbf{c)}};
        \node[rotate=90,text width=1.5cm] at (-1.4,12.8){Layer 1};
        \node[rotate=90,text width=1.5cm] at (-1.4,11.2){Layer 2};
        \node[rotate=90,text width=1.5cm] at (-1.4,9.6){Layer 3};

        \node[text width=2.0cm, align=center] at (-6.5,12.5){Colored seg. map};
        \node[text width=1.5cm, align=center] at (-4.5,14.0){One-hot vectors};
        \draw[thick,-T] (-3.5,11)--(-2,12.5);
        \draw[thick,-T] (-3.5,11)--(-2,11.0);
        \draw[thick,-T] (-3.5,11)--(-2,9.5);
        \node[rotate=45] at (-2.8,12.5){Smoothing};
        \node at (1.4,13.9){Ground Truth Maps};
        \node at (1.4,11.0){\includegraphics[width=5cm]{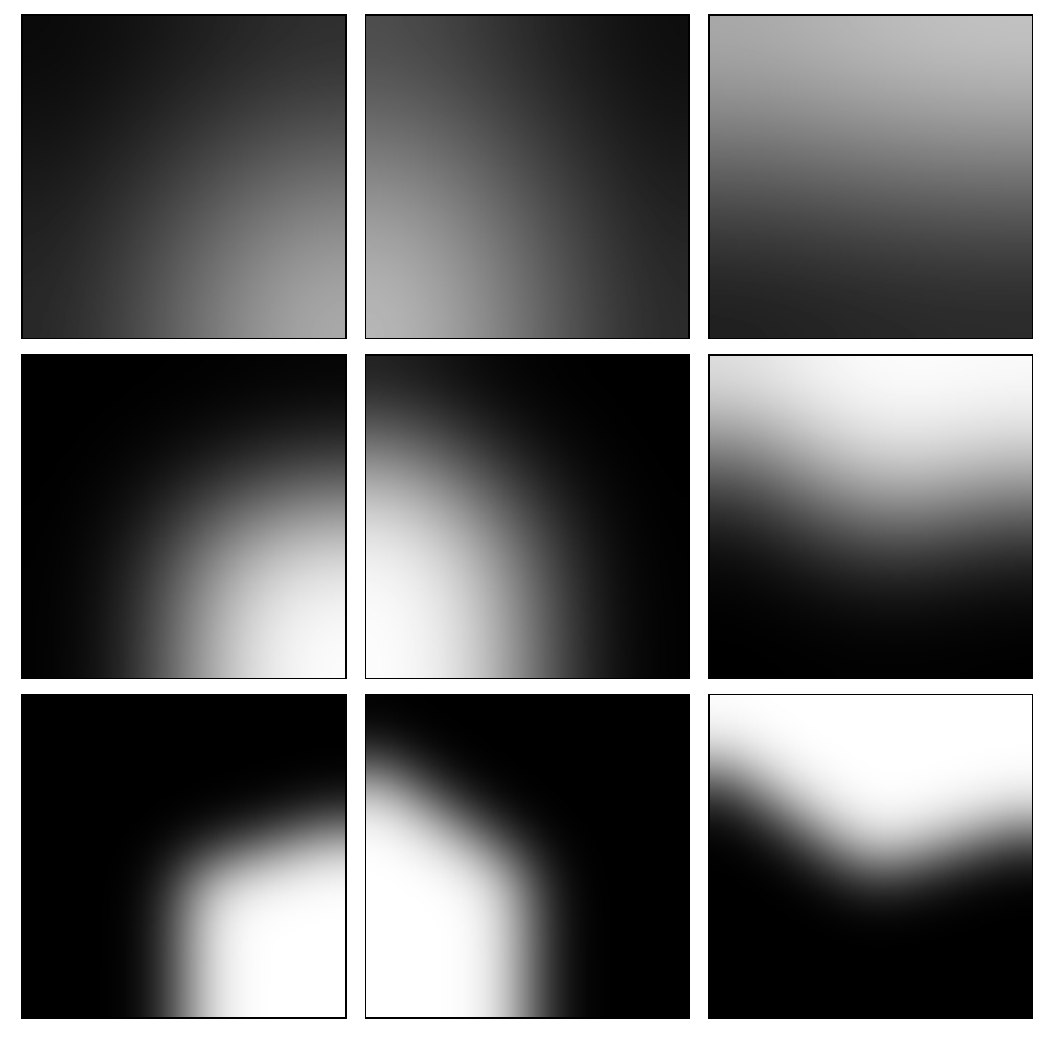}};
        \node at (4.85,13){Sampling};
        \draw[thick,-T] (4.1,12.6)--(5.5,12.6);
        \draw[thick,-T] (4.1,11.0)--(5.5,11.0);
        \draw[thick,-T] (4.1,9.4)--(5.5,9.4);

        \node at (6.6,14){Features};
        \node at (-3.0,8.0){Fitted Maps};
        \node at (3.0,8.0){Fitted Maps Indep.};

        \node at (-6.5,11.0){\includegraphics[height=1.82cm]{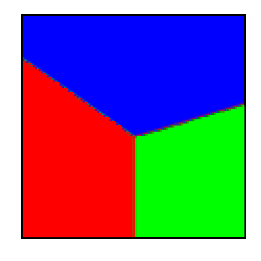}};
        \node at (-4.5,11.0){\includegraphics[height=5cm]{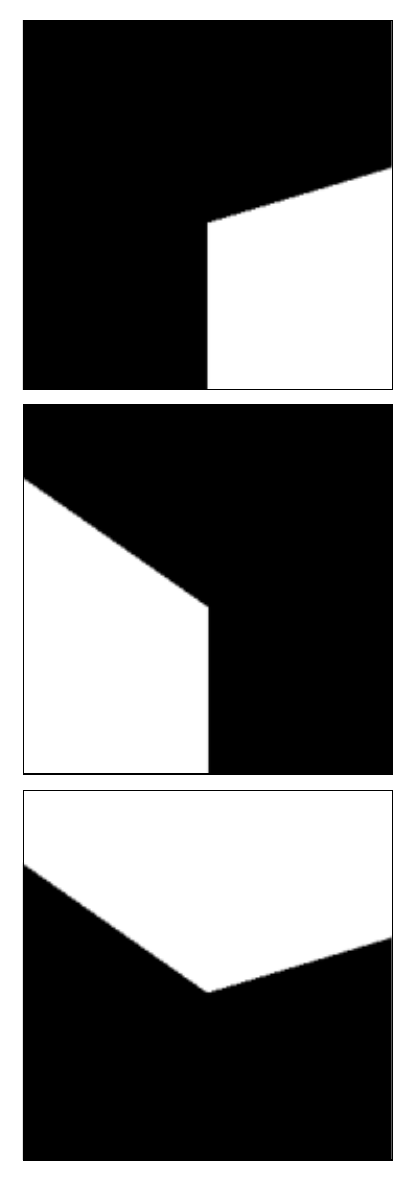}};

        \node at (6.6,11.0){\includegraphics[height=5cm]{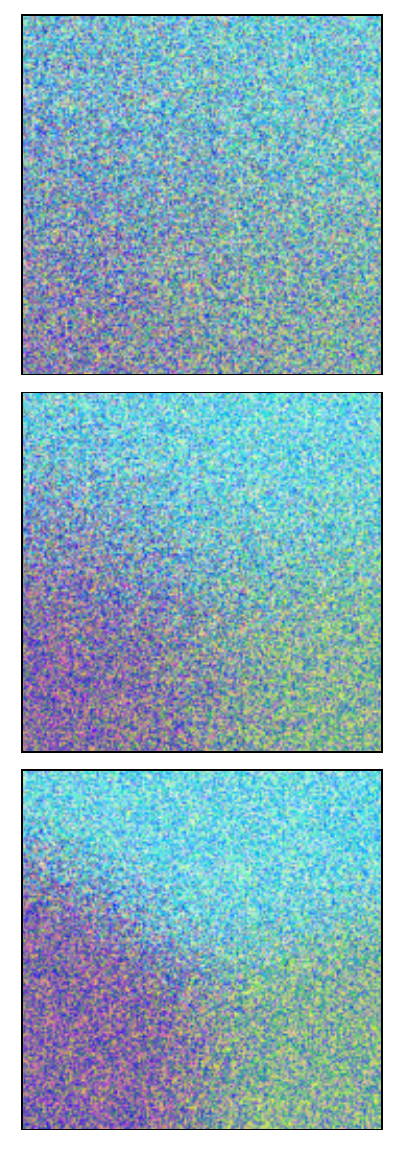}};
        \node at (-3.0,5.3){\includegraphics[width=5cm]{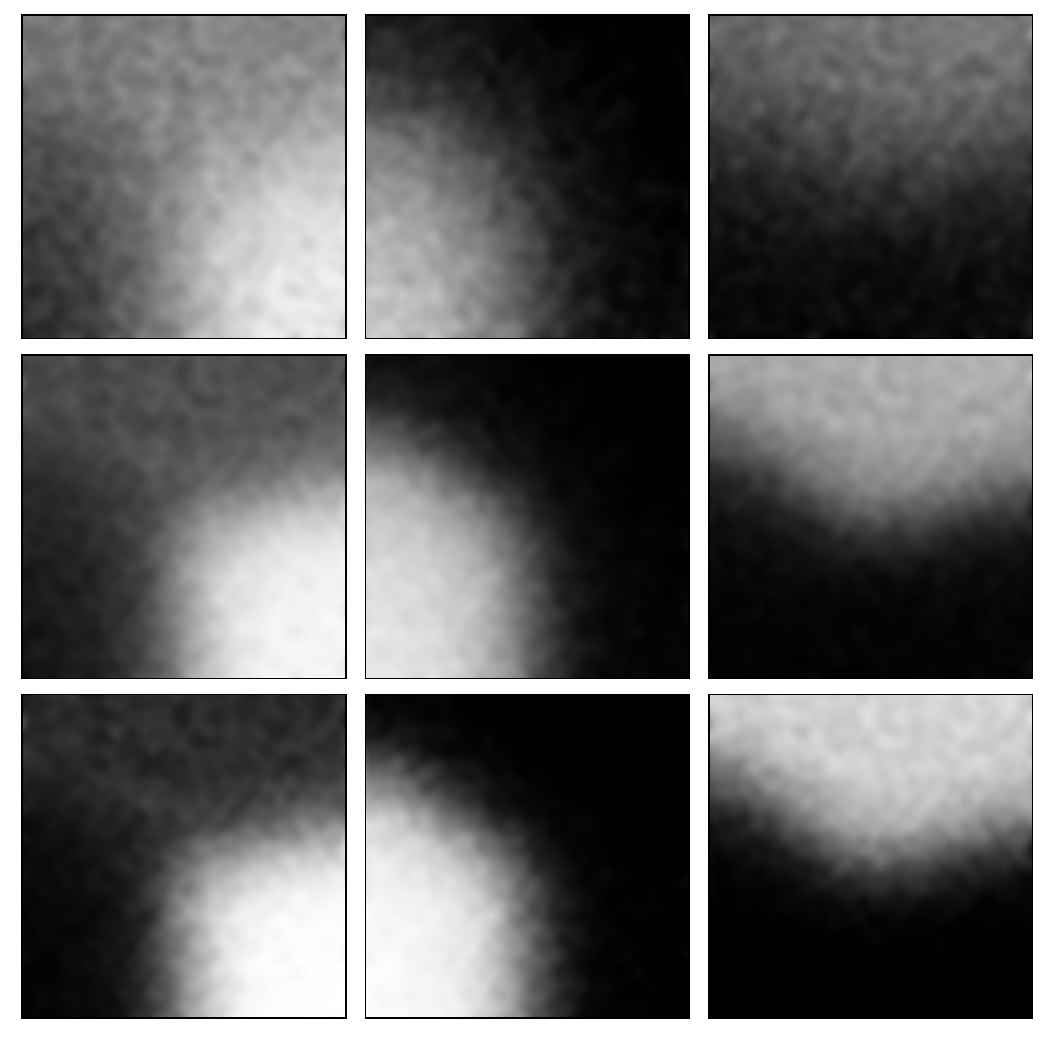}};
        \node at (3.0,5.3){\includegraphics[width=5cm]{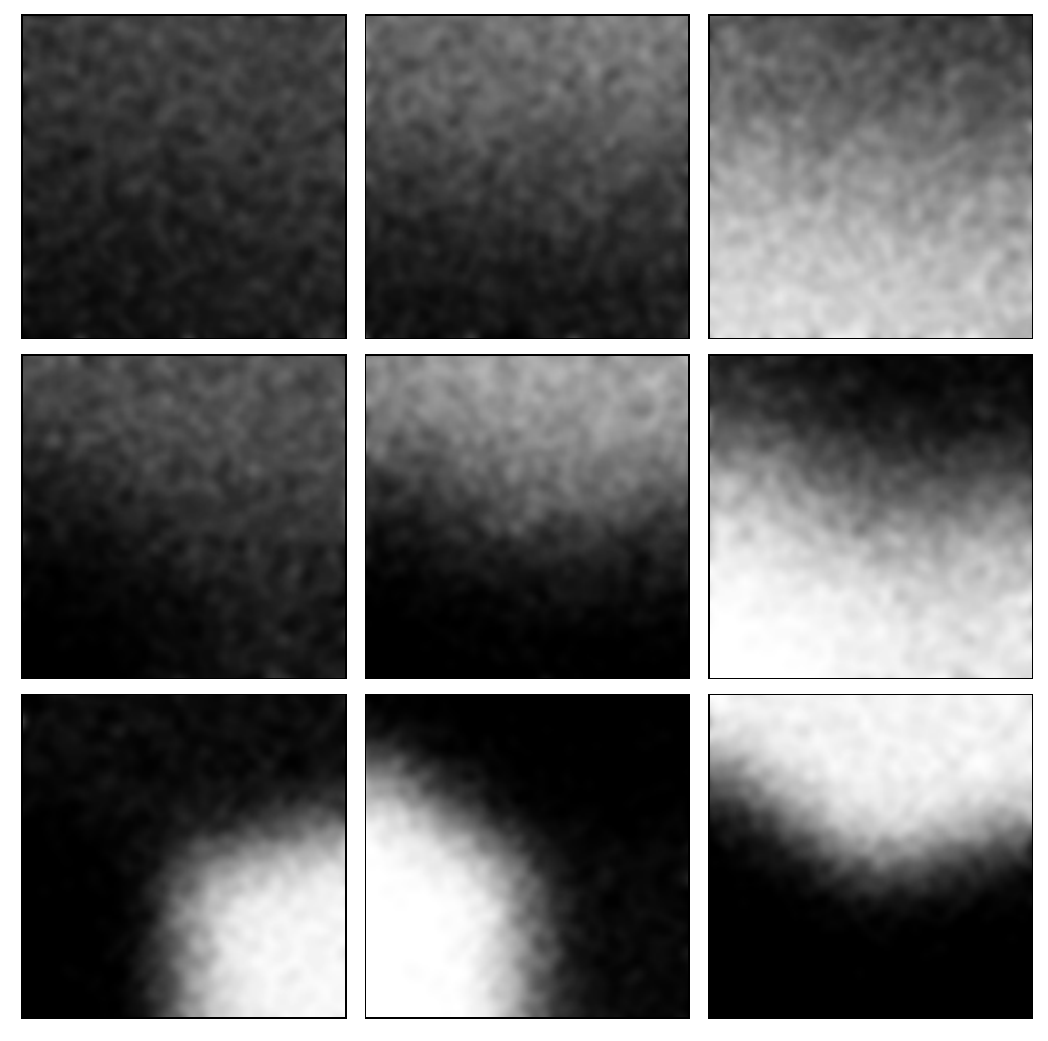}};
    \end{tikzpicture}
    \vspace{-1mm}
    \caption{Validation of hierarchical mixtures. \textbf{a)} Pipeline to generate the synthetic hierarchical data. Example synthetic images have three feature channels (displayed as RGB images on the right) and three components (columns of Ground Truth Maps), at three different layers (rows of Ground Truth Maps). \textbf{b)} Maps recovered by our hierarchical multi-layer model (Figure~\ref{fig:modified-mixture-multi}). Same conventions as in \textbf{a)}.  \textbf{c)} Maps recovered by applying our single mixture model independently at each layer. Same conventions as in \textbf{a)}. Figure is best seen in color.}
    \label{fig:results-fit-multi-mixture}
\end{figure}

\subsection{Application to Natural Image Segmentation}

To test our algorithms on natural images, we considered unsupervised segmentation of natural images. This is a classical application of mixture model, and particularly relevant for our framework, given that natural images have both spatial and hierarchical structure, as demonstrated by the success of DNN models on many visual tasks. We therefore applied our algorithms to natural images from the BSD500 dataset~\cite{arbelaez2011contour}, and compared the results between different variants of our model and to the segmentation maps of human observers.

\subsubsection{Single Mixture}

\paragraph{Implementation details} As in the synthetic image experiments, we used RGB colors as features. In addition to GMM, we ran our algorithm using Student-t Mixture Models (SMM), that is using Student-t distributions to describe the data distribution of each mixture component. The SMM is more robust to outliers, and the comparison with the GMM allowed us to assess the importance of such robustness for segmentation performance~\cite{coencagli2009stat,coencagli2012surround,wainwright2000scale}. For both GMM and SMM, we also compared the results with and without using our smoothing priors, to study their relevance for segmentation. As the number $K$ of components is unknown, we also ran our algorithm with $K=3$, $6$ and $9$, and studied the results separately. As for synthetic data, the functions  ${u}_{n,k}$ are 2-D Gaussians, with width $\sigma=2.75$. \revisedA{This choice corresponds to spatial smoothing, and is motivated by the evidence from previous algorithms that spatial smoothing improves image segmentation. In addition, it allows us to compare directly the performance of our algorithm to~\cite{nikou2010bayesian} and~\cite{sun2017location}.} To obtain segmentation maps directly comparable with those measured in humans, we used a maximum a posteriori (MAP) approach: we assign each pixel to the component with the highest probability, and therefore discard the probabilistic information.

\begin{wrapfigure}[25]{r}{0.54\textwidth}
    \centering
    \includegraphics[width=8cm]{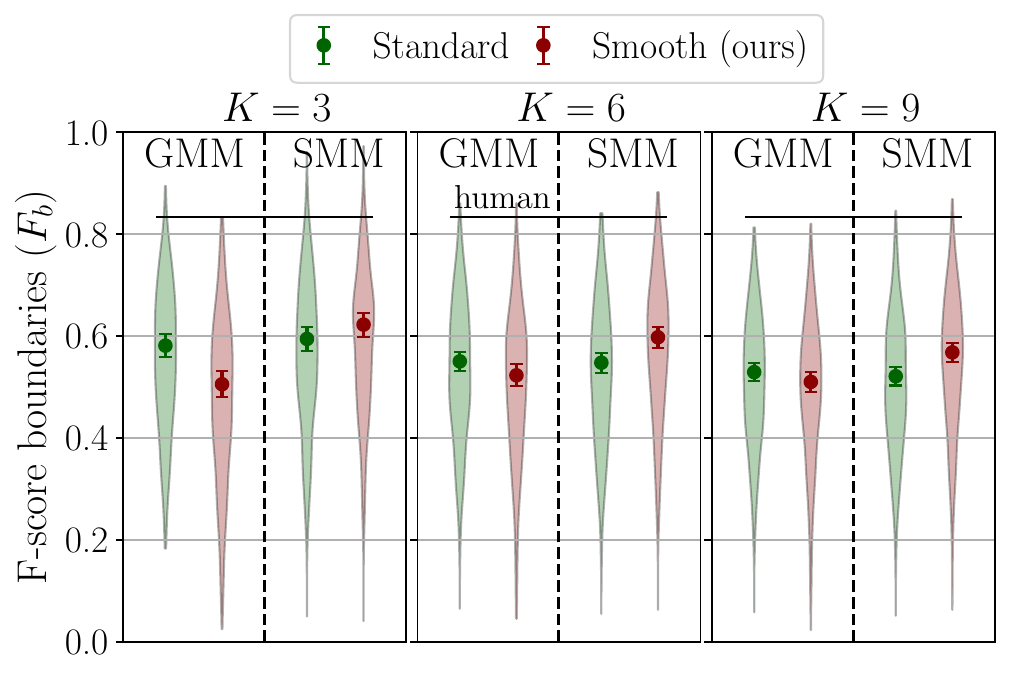}\\
    \includegraphics[width=8cm]{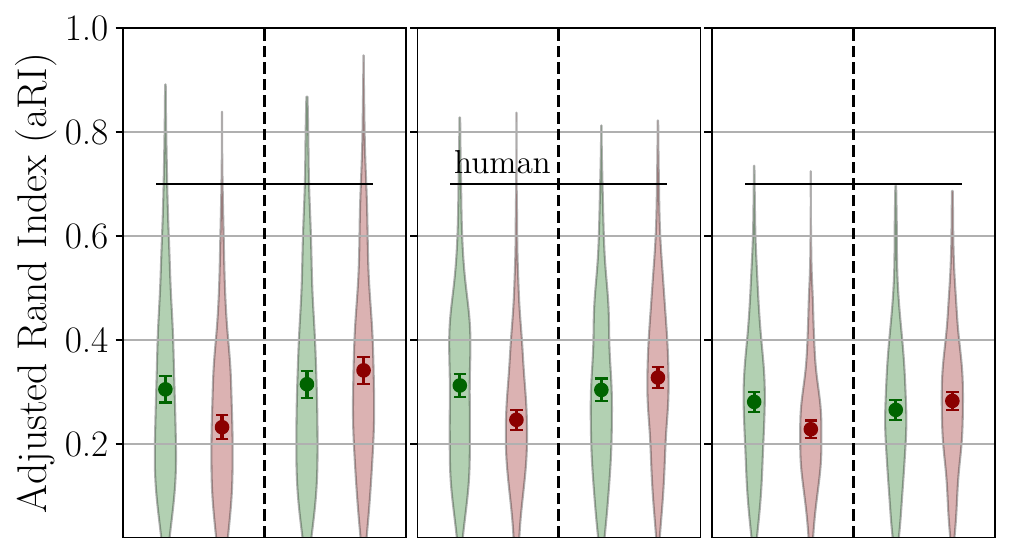}
    \caption{Performance of single mixture models trained with RGB color features on the BSD 500 for the two scores defined in text. Error bars indicate 3 times the standard error of the mean. Shaded areas represent scores density. Figure is best seen in color.}
    \label{fig:qt-single-results}
\end{wrapfigure}

\paragraph{Quantitative Results} To quantify the performance of our algorithms, we used two widely-adopted scores, that measure the similarity between the segmentation maps of humans and of the algorithms: the adjusted Rand Index (aRI)~\cite{hubert1985comparing} and the F-score for boundaries ($F_b$)~\cite{pont2013measures}. The aRI measures the overlap between regions in the two segmentation maps, whereas $F_b$ measures the consistency of boundaries between regions. The results are reported in Figure~\ref{fig:qt-single-results}. Overall performances of mixture models are much lower than those of humans (which is assessed by quantifying consistency across different observers), and generally the best performances are achieved using the SMM and smoothing. Interestingly, for all tested $K$, our smoothing improves both scores for SMM while it impairs both scores for GMM. The reason is that the smoothing dramatically affects the parameters learned in Gaussian components while the robustness of the Student's components prevents this effect. Furthermore, increasing the number of components decreases the average scores in general, although the best $K$ varies substantially across images. The average decrease in performance at larger $K$ might be due to an increase in the detection of false contours.

\begin{wrapfigure}[27]{r}{0.54\textwidth}
    \centering
    \includegraphics[width=8cm]{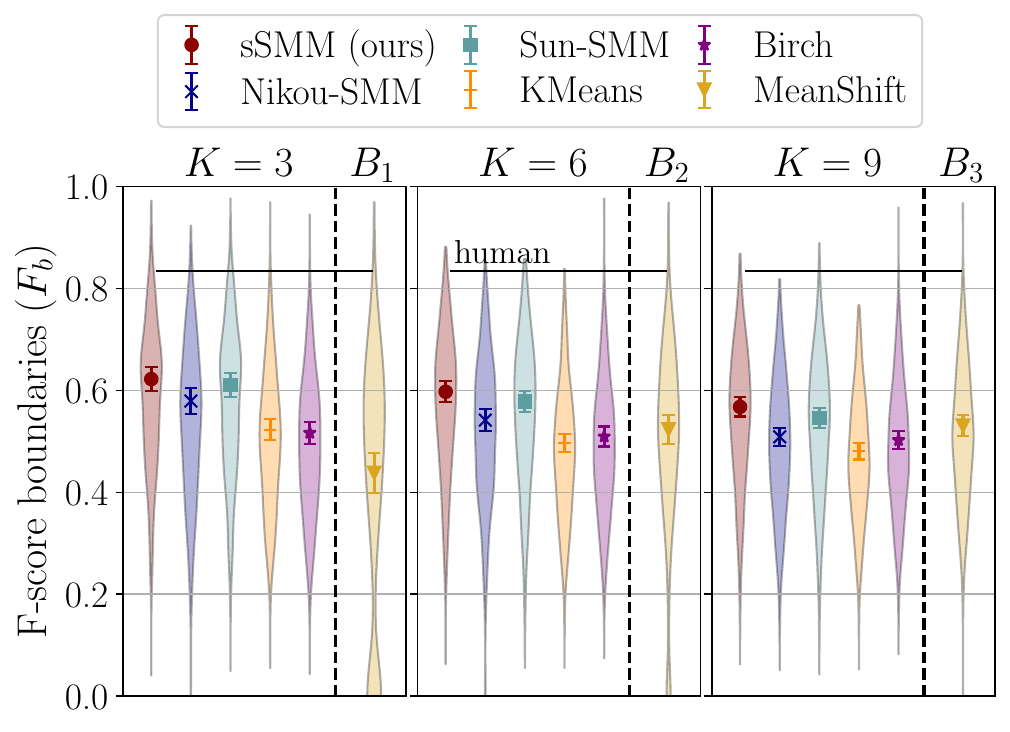}\\
    \includegraphics[width=8cm]{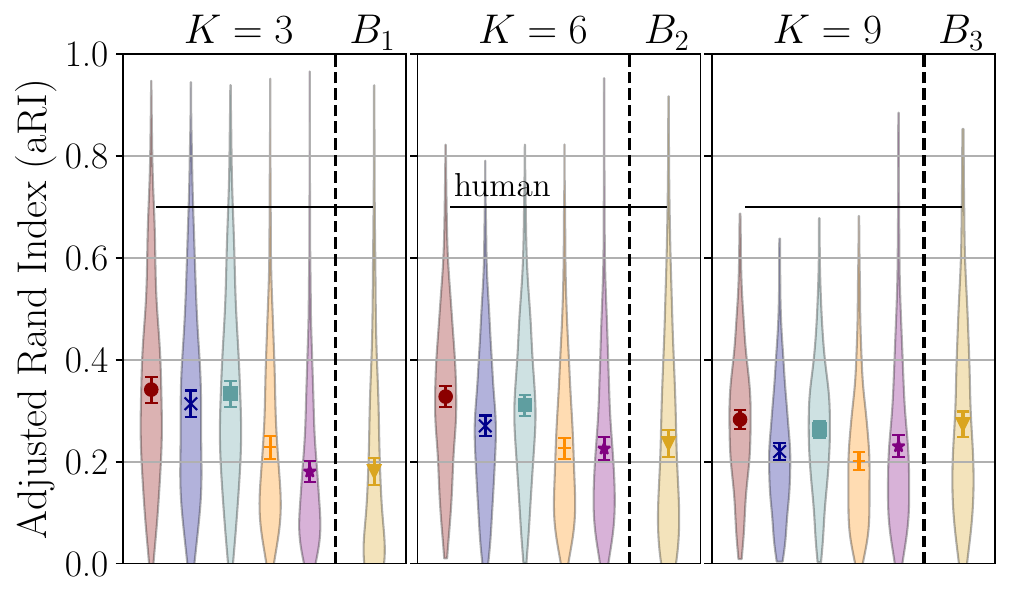}
    \vspace{-8mm}
    \caption{\revisedA{Comparison of our best model with other unsupervised learning models. All algorithms are trained with RGB color features on the BSD 500, and evaluated on the two scores defined in text. Error bars indicate 3 times the standard error of the mean. Shaded areas represent scores density. Figure is best seen in color.}}
    \label{fig:qt-single-results-comp}
\end{wrapfigure}

\paragraph{\revisedA{Comparison with Other Unsupervised Learning Algorithms}} \revisedA{We compare our algorithm to the following classical unsupervised learning algorithms: KMeans~\citep{steinhaus1956kmeans}, Birch~\citep{zhang1996birch} and MeanShift~\citep{comaniciu2002mean}.
In addition, we also compare to DCM-SVFMM of Nikou~\etal~\cite{nikou2010bayesian} and LDDP of Sun~\etal~\cite{sun2017location}, the two smoothing models that we presented in Section \ref{sec:intro}. All those algorithms rely on a parameter which controls the number of clusters, either directly or using the bandwidth for MeanShift (labeled $B_1$, $B_2$ and $B_3$ in the Figure). The quantitative results are shown in Figure~\ref{fig:qt-single-results-comp}. For both scores and when $K=3$ or $6$, the three smoothing-based models (\cite{nikou2010bayesian}, \cite{sun2017location} and ours) perform better than the three other models. For $K=9$, only our model performs marginally better than others. Among the three smoothing models, ours performs best, followed by \cite{sun2017location} and then \cite{nikou2010bayesian}. However, the difference is sometimes marginal (overlapping error bars for $K=3$ in both scores).
}

\begin{figure}
    \centering
    \begin{tikzpicture}
        \draw[step=1.0,white,thin] (-7.5,9) grid (7,-9);
        \node at (-3.5,8.75){Original Image};
        \node at (-5.9,6.825){$K=3$};
        \node at (-3.5,6.825){$K=6$};
        \node at (-1.2,6.825){$K=9$};
        \node[rotate=90] at (-7.25,5.825){sSMM};
        \node[rotate=90] at (-7.25,4.3){SMM};
        \node[rotate=90] at (-7.25,2.8){sGMM};
        \node[rotate=90] at (-7.25,1.25){GMM};

        \node at (3.5,8.75){Original Image};
        \node at (5.9,6.825){$K=9$};
        \node at (3.5,6.825){$K=6$};
        \node at (1.2,6.825){$K=3$};

        \node at (-3.5,-0.25){Original Image};
        \node at (-5.9,-2.175){$K=3$};
        \node at (-3.5,-2.175){$K=6$};
        \node at (-1.2,-2.175){$K=9$};
        \node[rotate=90] at (-7.25,-3.125){sSMM};
        \node[rotate=90] at (-7.25,-4.75){SMM};
        \node[rotate=90] at (-7.25,-6.25){sGMM};
        \node[rotate=90] at (-7.25,-7.75){GMM};

        \node at (3.5,-0.25){Original Image};
        \node at (1.2,-2.175){$K=3$};
        \node at (3.5,-2.175){$K=6$};
        \node at (5.9,-2.175){$K=9$};

        \node at (-3.5,4.5){\includegraphics[width=7cm]{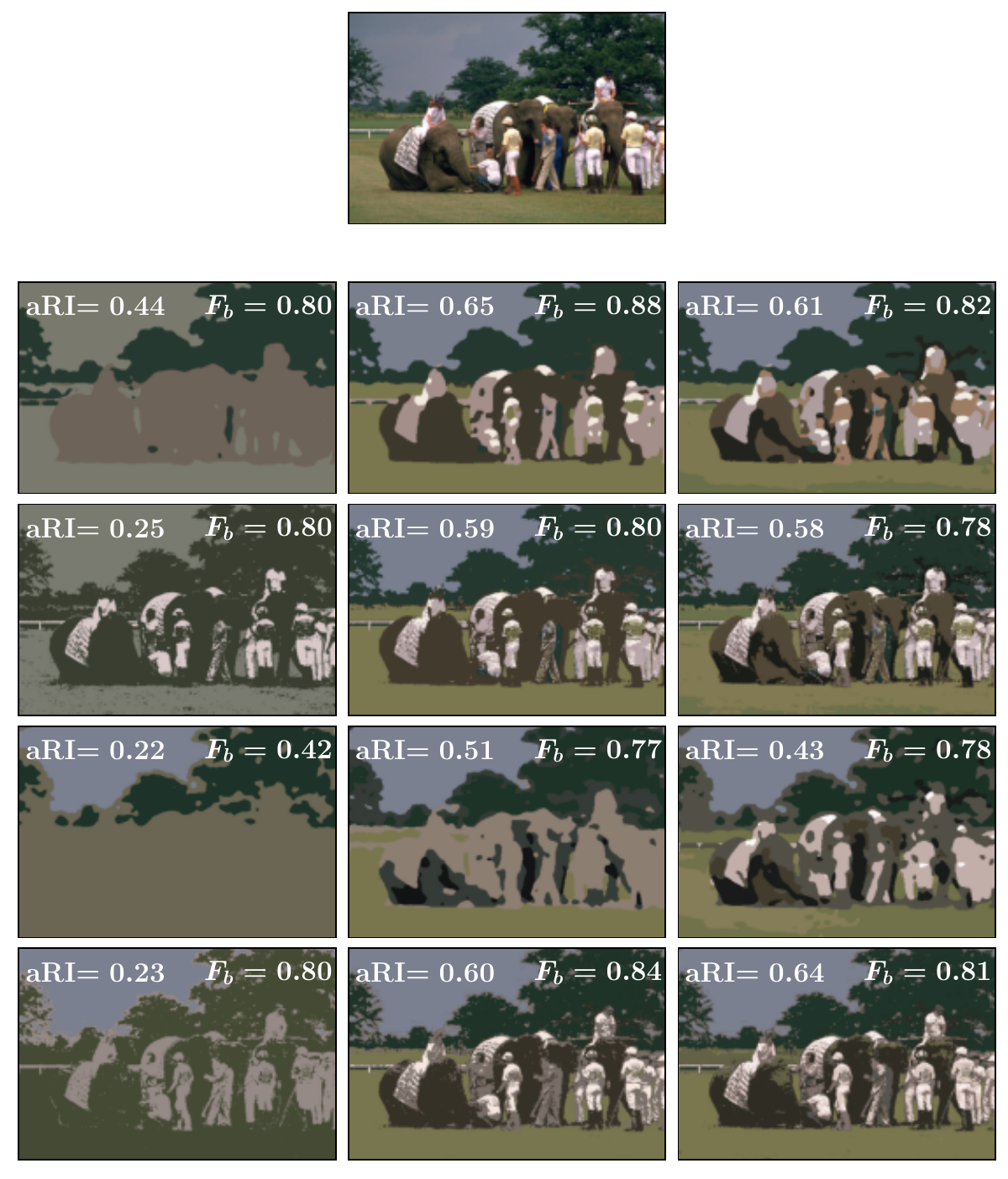}};
        \node at (3.5,4.5){\includegraphics[width=7cm]{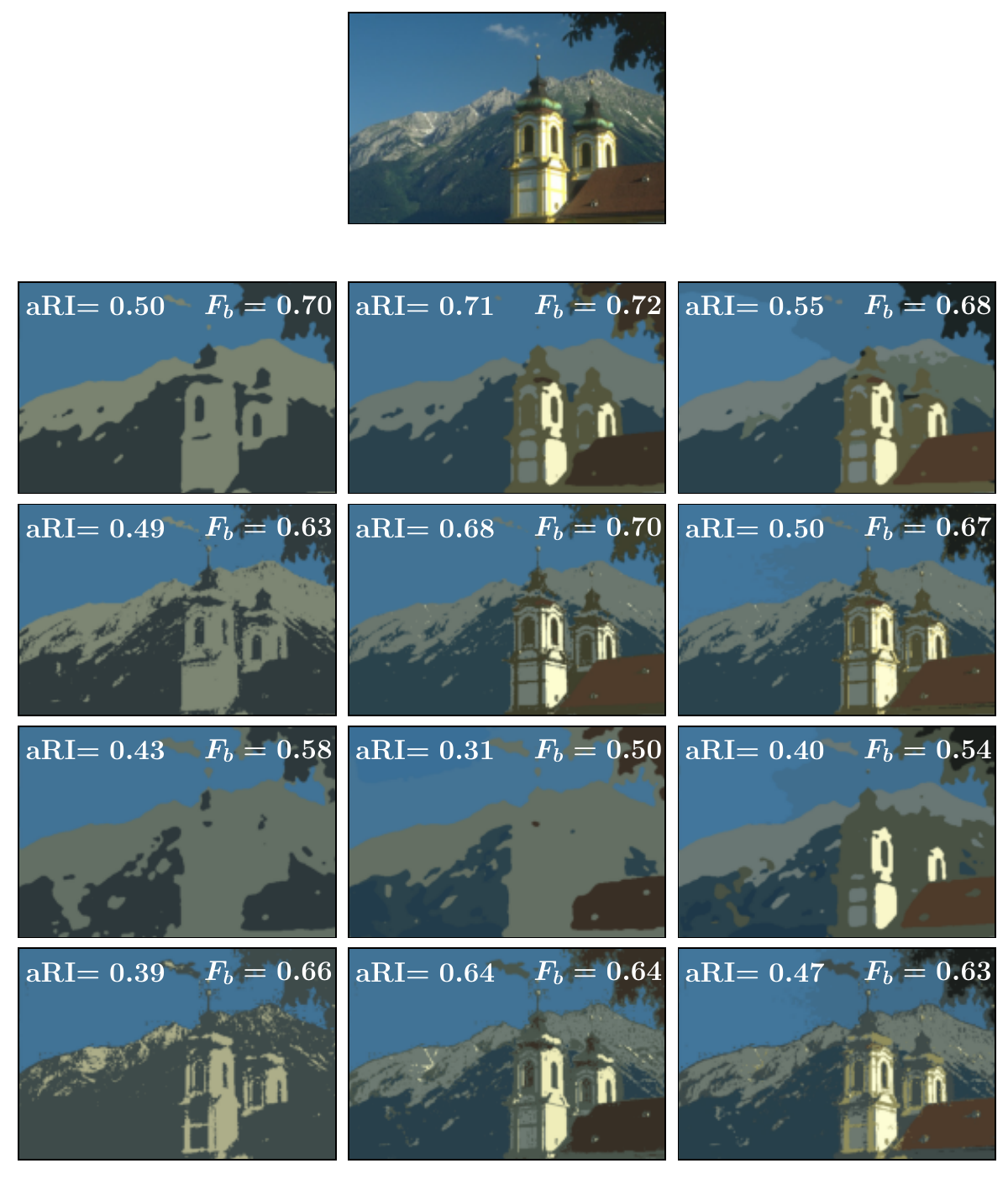}};
        \node at (-3.5,-4.5){\includegraphics[width=7cm]{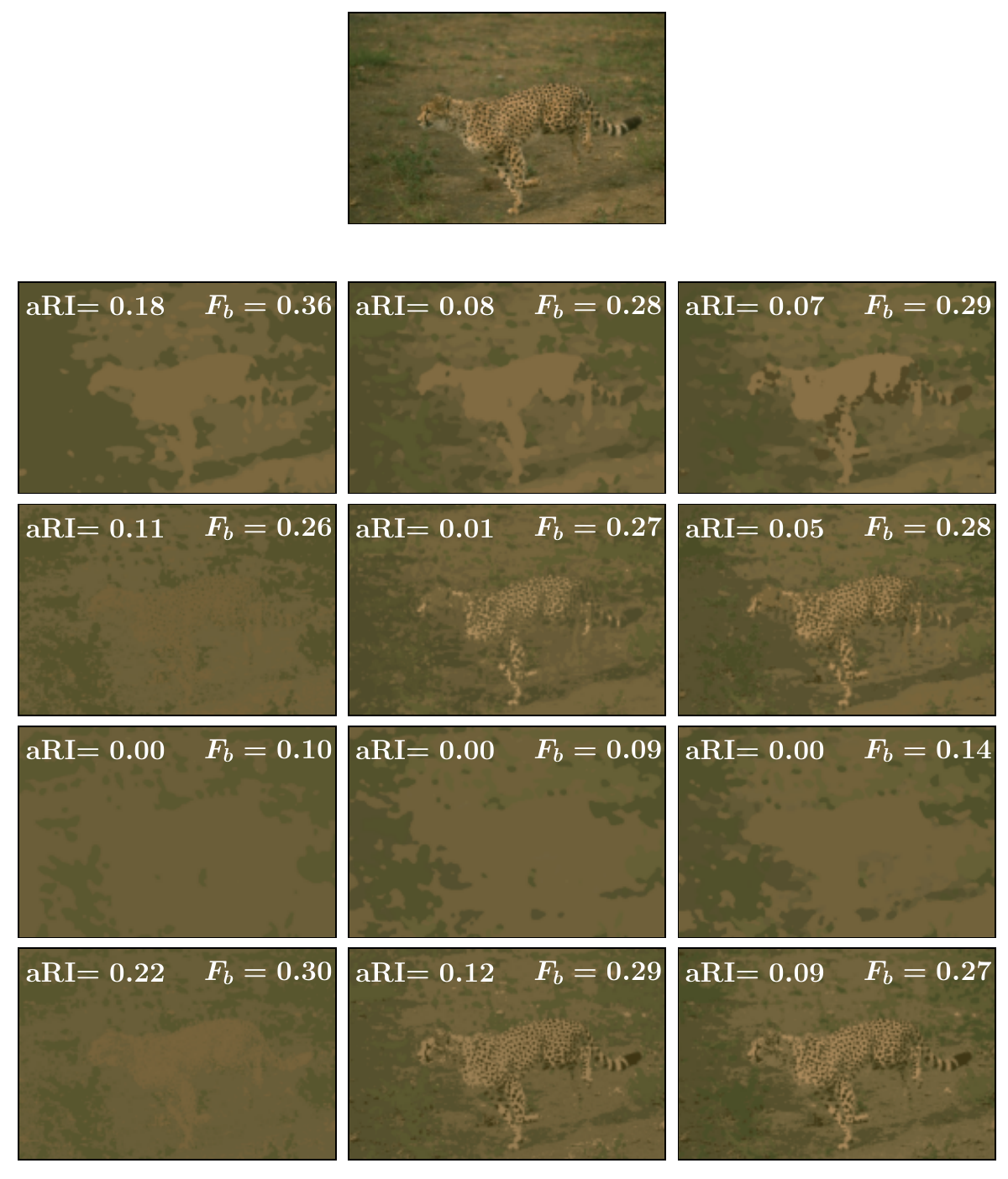}};
        \node at (3.5,-4.5){\includegraphics[width=7cm]{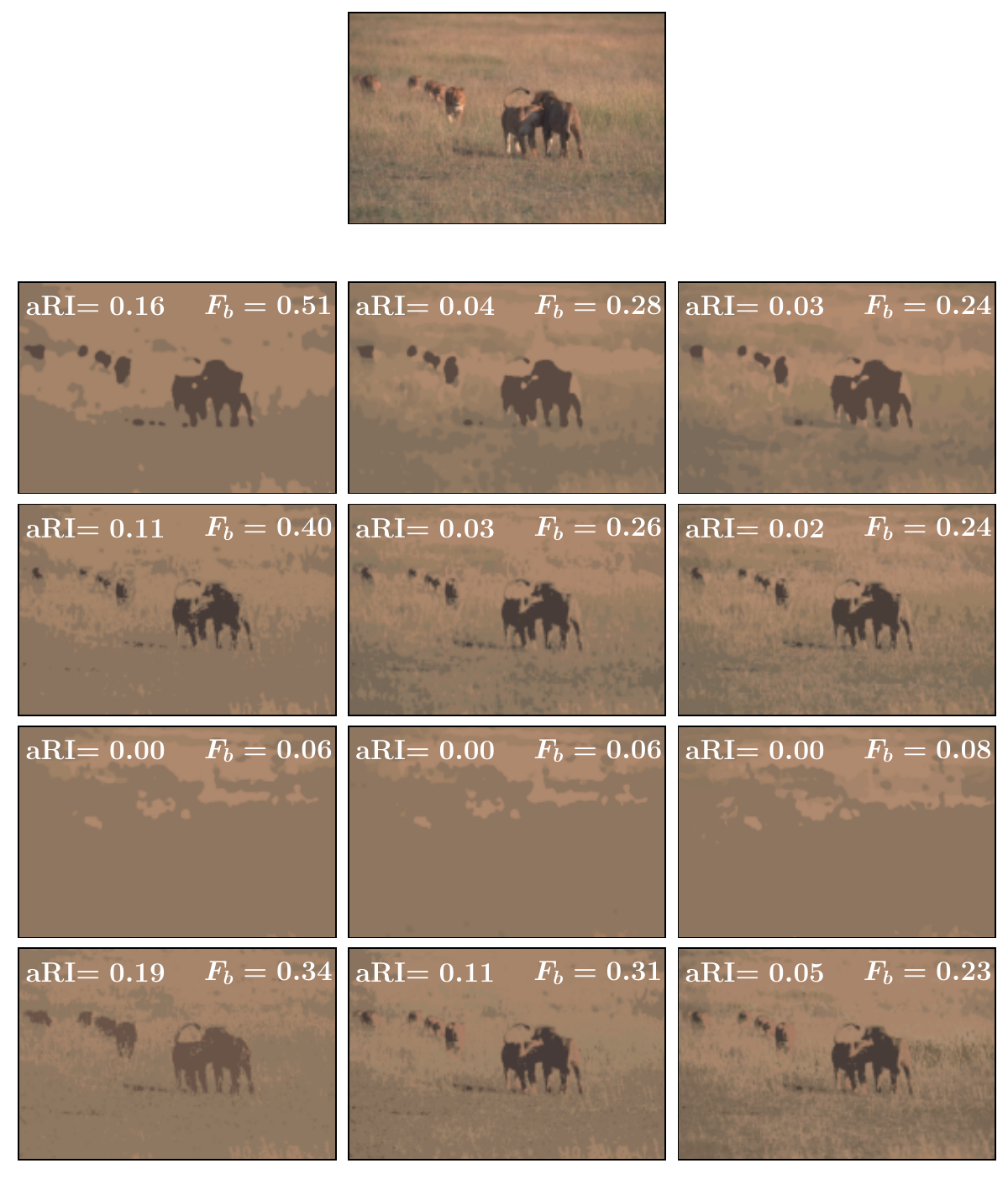}};
    \end{tikzpicture}
    \caption{Segmentation of natural images using mixture models with (sGMM/sSMM) and without (GMM/SMM) smoothing. Top: high-aRI images. Bottom: low-aRI images. We display the area corresponding to each component with the average color of the original image inside that area. For each image, the aRI and $F_b$ scores are shown in white. Figure is best seen in color.}
    \label{fig:ql-single-results}
\end{figure}

\paragraph{Qualitative Results} In Figure~\ref{fig:ql-single-results}, we show four segmentation examples among which two have high aRI scores (top) and two have low aRI scores (bottom). The main qualitative difference is that images with high score have sharp contours delimiting weakly textured areas, and different areas have largely different color, whereas low score images have blurry contours delimiting rough textured areas. Similar to what we have shown in the synthetic data, our smoothing enforces contiguity of component areas. The examples of Figure~\ref{fig:ql-single-results} also illustrate some aspects of our quantitative results. \revisedB{First, increasing $K$ does not improve scores on average. For instance, for the top right image, when we train mixture models with $K=9$ components, we observe false contours (in the sky and mountain) and therefore the emergence of superfluous areas, which impairs both scores. Second, the four selected examples illustrate how smoothing improves the scores for SMM while it diminishes the scores for GMM.}
\revisedB{For instance, for the bottom-left image, we observe that the leopard is well-delimited by the GMM, whereas the smoothing mixes the leopard with the background.} This is not the case for SMM with smoothing, thanks to the robustness of the Student-t distribution.

\subsubsection{Multiple Mixtures}

\paragraph{Implementation details} For the hierarchical observations $(x_n^{(h)})_{n,h}$, we used image features extracted by the pre-trained deep network VGG 19~\cite{simonyan2014very}. We compared Gaussian (GMM) and Student-t (SMM) mixture components. The Student-t distribution captures the sparse, heavy-tailed behavior of both low-level features (\eg{} wavelet coefficients~\cite{wainwright2000scale,vacher2020texture}) and higher-level features extracted by DNN~\cite{sanchez2019normalization}. Therefore, comparing GMM to SMM allows us to study the importance of modeling accurately the component distributions, for segmentation performance. To assess the relative importance of integrating information across spatial and hierarchical dimensions, we implemented two models. Model \textbf{a} used a nearest-neighbor hierarchical structure, similar to Figure~\ref{fig:modified-mixture-multi}. Therefore, in this model, each layer integrated information spatially as well as from one layer below and one above. Because model \textbf{a} resulted in a separate (though not independent) segmentation map per feature layer, \revisedA{we compared two approaches to extract a single segmentation map for performance evaluation and for model comparison: (i) we computed the product of all the probability maps at all layers, and then computed the MAP segmentation map, strategy that we call Model a; (ii) we computed the MAP segmentation map from the probabilities of the first layer only, strategy that we call Model a / 1st layer.} Lastly, we also considered a second model (model \textbf{b}), in which the observations at all layers shared a single segmentation map, \ie{} a single set of class labels $(C_n)_n$. Therefore, in model \textbf{b}, the prior parameters and corresponding mixing probabilities were influenced by the observations at all layers, with uncertainty-weighting. Again, by choosing the function $u_{n,k}^{(h)}$ appropriately, model \textbf{a} has the update rule given by Equation~\eqref{eq:prior-update-a} whereas model \textbf{b} has the following update rule
\eql{\label{eq:prior-update-b}
p_{n,k}^{(1,t+1)} = \frac{\sum_{h=1}^H \prod_{i\neq h} {s_n^{(i,t)}}^2 m_{n,k}^{(h,t)}}{\sum_{h=1}^H \prod_{i\neq h} {s_n^{(i,t)}}^2},
}
where $m_{n,k}^{(h,t)}$ and ${s_n^{(h,t)}}^2$ are respectively the local mean and variance of the posterior maps at layer $h$, defined in Equation~\eqref{eq:loch} and~\eqref{eq:scale} respectively; $\tau_{k}^{(h,t)}: l_n \mapsto \tau_{n,k}^{(h,t)}$ is the posterior map at layer $h$; and $G^{(h)}$ is a 2-D Gaussian kernel with width $\sigma^{(h)}$. The two models~\textbf{a} and \textbf{b} are fitted using Algorithm~\ref{alg:EMlayers}, with respectively the modified update rule~\eqref{eq:prior-update-a} and \eqref{eq:prior-update-b}. We used the following values of $\sigma^{(h)}$: 4.25, 4.25, 3.25, 3.25, 2.25, 2.25, 2.25, 2.25, 0.75, $\dots$, 0.75 for both models.
\begin{wrapfigure}[26]{r}{0.54\textwidth}
    \centering
    \includegraphics[width=8cm]{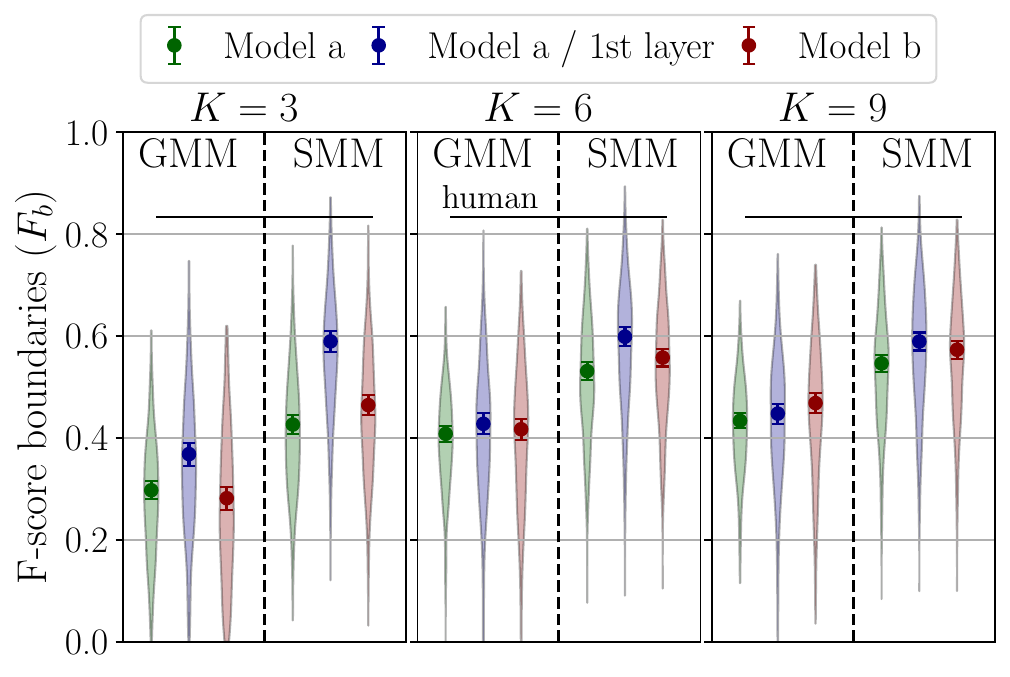}\\
    \includegraphics[width=8cm]{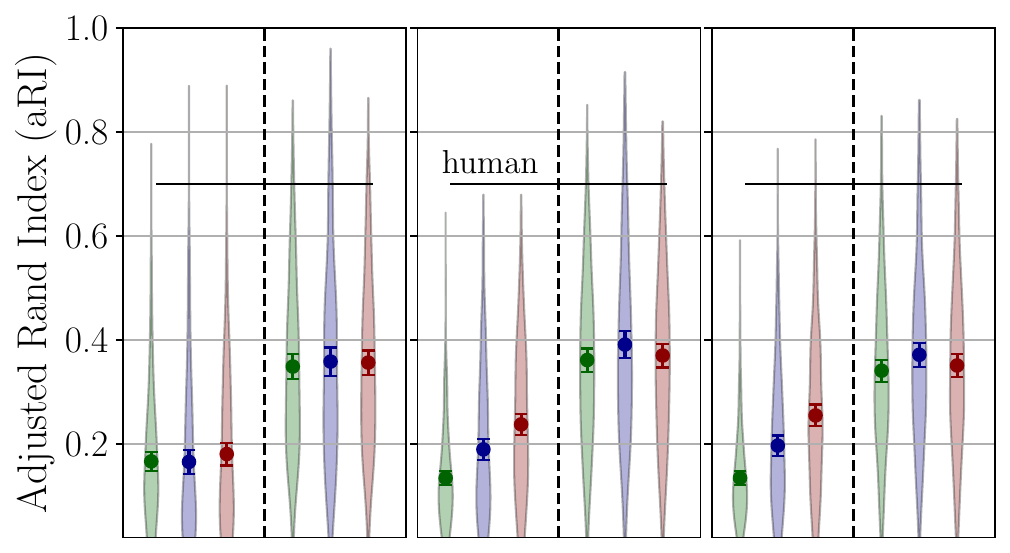}
    \caption{Performance of model \textbf{a} and \textbf{b} trained with VGG 19 features on the BSD 500 for the two scores defined in text. Error bars and shaded areas: same convention as in Figure \ref{fig:ql-single-results}. For each image, the aRI and $F_b$ scores are shown in white. Figure is best seen in color.}
    \label{fig:qt-multiple-results}
\end{wrapfigure}

\noindent Importantly, the modified update rules implement uncertainty-based integration: the local mean of each layer is weighted by the local variances of the other layers, such that a layer with higher uncertainty (large variance) will contribute relatively less to the update of the other layers. We adapted our implementation of the models to the structure of VGG features. First, the number of pixels (\ie{} the number of samples) is not same in the different layers, therefore we up-sampled the posterior probability maps $\tau_{k}^{(h,t)}$ using nearest neighbor interpolation before convolution with $G^{(h)}$ when it was necessary. Second, the decreasing number of samples and the increasing dimension of features along the depth of the network often caused numerical issues, therefore we reduced the dimension at each layer using Principal Component Analysis (PCA), including only the dimensions needed to capture $95\%$ of the variance. Third, the first layer of the deep network is a linear transform of the input image, in contrast with all subsequent layers, therefore we added the features of the first layer to all subsequent layers (using average pooling when necessary).

\begin{figure}[b]
    \centering
    \includegraphics[width=6cm]{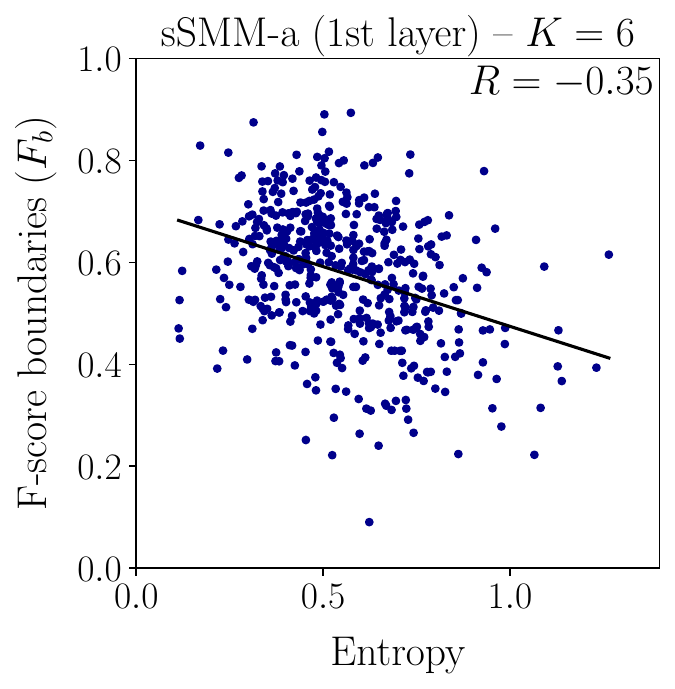}
    \includegraphics[width=6cm]{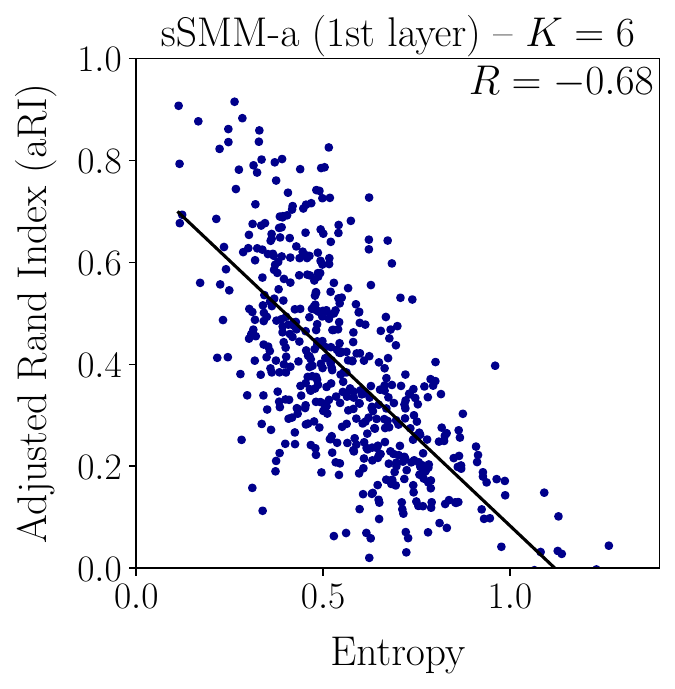}
    \caption{Negative correlation between both scores and entropy of model \textbf{a}/1st layer when using Student-t mixtures.}
    \label{fig:score-vs-entropy}
\end{figure}

\paragraph{Quantitative Results} The F-score and aRI of model \textbf{a}, model \textbf{a}/1st layer only, and model \textbf{b} are reported in Figure~\ref{fig:qt-multiple-results}. The use of VGG features impairs contour detection in comparison to the use of RGB features, for the GMMs and the SMM without smoothing. Indeed the F-score is much lower for all model variants except for model \textbf{a}/1st layer and model \textbf{b} both with SMM (Figure~\ref{fig:qt-multiple-results}, top \vs{} Figure~\ref{fig:qt-single-results}, top). Increasing the number of component $K$ tends to recover the performance on the F-scores, for the models that are impaired by VGG features. The impairment of contour detection is due to the reduced resolution of the maps at deeper layer, which influences the combined segmentation map, but much less the map obtained with the 1st layer only. In contrast, the use of VGG features improves the aRI of SMM while impairing the aRI of GMM, in comparison with RGB features (Figure~\ref{fig:qt-multiple-results}, bottom \vs{} Figure~\ref{fig:qt-single-results}, bottom). Yet, the use of combined maps does not improve the aRI score of the 1st layer segmentation maps. Increasing the number of components $K$ has almost no effect on average performances, although, as for RGB, the best $K$ varies substantially across images. Overall performances of model \textbf{b} are higher than those of model \textbf{a} but are lower than those of the model \textbf{a}/1st layer. Lastly, as for RGB, the performance of both models is much lower than humans. We show in Figure~\ref{fig:score-vs-entropy} that both scores are negatively correlated with the level of uncertainty captured by the Student-t mixture model. Such correlation is interesting because it shows that resolving high uncertainty areas using additional low uncertainty features can improve segmentation quality. In addition, this could explain the marginal superiority of model \textbf{b} over the combination of probabilities of model \textbf{a}. Indeed model \textbf{b} weights low uncertainty layer maps more which should lead to higher scores. This observation does not hold for Gaussian mixture (see supplementary Figure~\ref{supp-fig:score-vs-entropy}).
\begin{figure}
    \centering
    \begin{tikzpicture}
        \draw[step=1.0,white,thin] (-7.5,9) grid (7,-14.5);
        \node at (-3.5,8.8){Original Image};
        \node at (-5.9,6.75){$K=3$};
        \node at (-3.5,6.75){$K=6$};
        \node at (-1.2,6.75){$K=9$};
        \node[rotate=90,font=\scriptsize] at (-7.25,5.65){sSMM-a1};
        \node[rotate=90,font=\scriptsize] at (-7.25,4.15){sSMM-a};
        \node[rotate=90,font=\scriptsize] at (-7.25,2.6){sSMM-b};
        \node[rotate=90,font=\scriptsize] at (-7.25,1.05){sGMM-a1};
        \node[rotate=90,font=\scriptsize] at (-7.25,-0.45){sGMM-a};
        \node[rotate=90,font=\scriptsize] at (-7.25,-2.0){sGMM-b};

        \node at (3.5,8.8){Original Image};
        \node at (5.9,6.75){$K=9$};
        \node at (3.5,6.75){$K=6$};
        \node at (1.2,6.75){$K=3$};

        \node at (-3.5,-2.95){Original Image};
        \node at (-5.9,-5.0){$K=3$};
        \node at (-3.5,-5.0){$K=6$};
        \node at (-1.2,-5.0){$K=9$};
        \node[rotate=90,font=\scriptsize] at (-7.25,-6.05){sSMM-a1};
        \node[rotate=90,font=\scriptsize] at (-7.25,-7.55){sSMM-a};
        \node[rotate=90,font=\scriptsize] at (-7.25,-9.15){sSMM-b};
        \node[rotate=90,font=\scriptsize] at (-7.25,-10.75){sGMM-a1};
        \node[rotate=90,font=\scriptsize] at (-7.25,-12.25){sGMM-a};
        \node[rotate=90,font=\scriptsize] at (-7.25,-13.75){sGMM-b};

        \node at (3.5,-2.95){Original Image};
        \node at (5.9,-5.0){$K=9$};
        \node at (3.5,-5.0){$K=6$};
        \node at (1.2,-5.0){$K=3$};

        \node at (-3.5,2.9){\includegraphics[width=7cm]{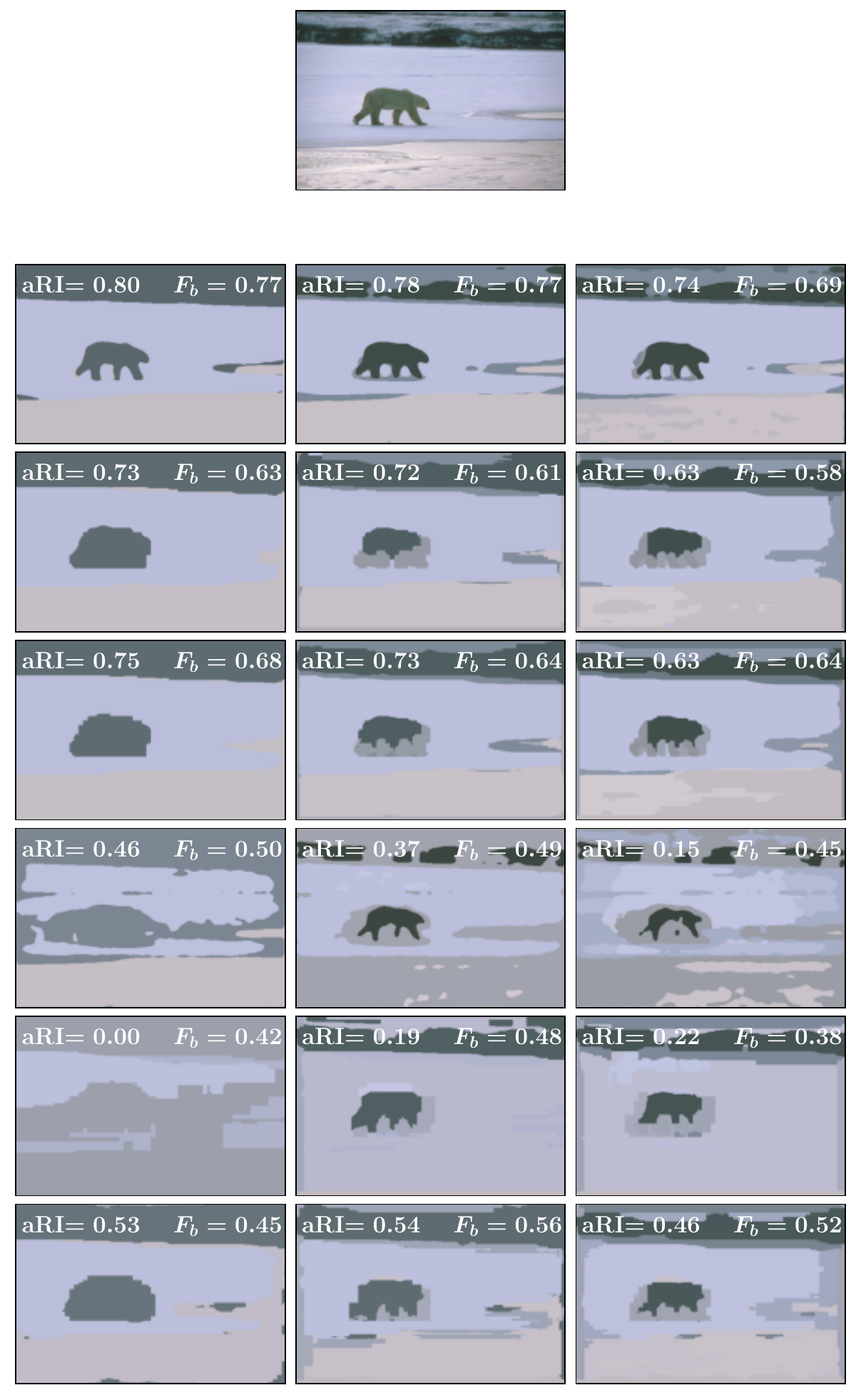}};
        \node at (3.5,2.9){\includegraphics[width=7cm]{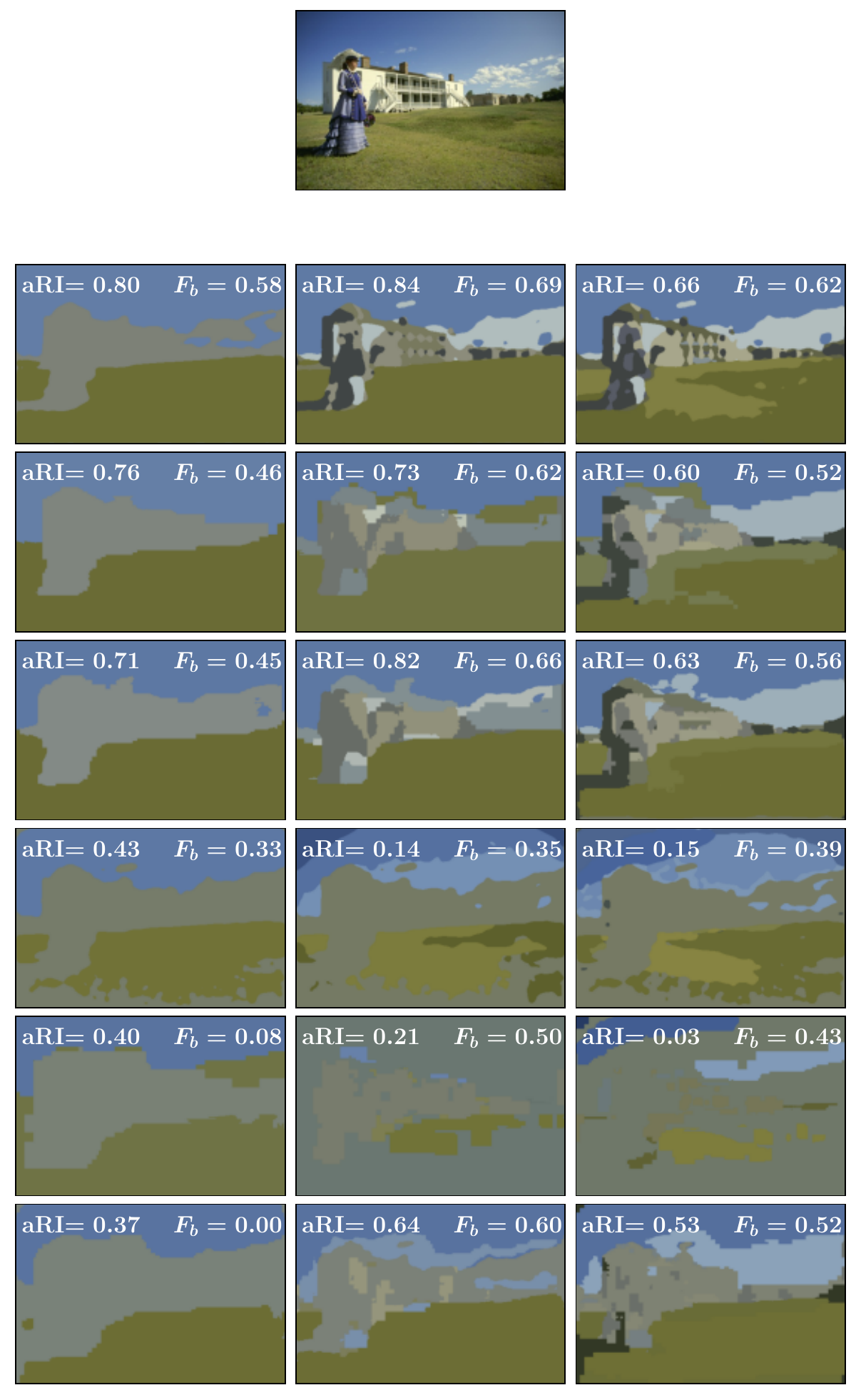}};
        \node at (-3.5,-8.85){\includegraphics[width=7cm]{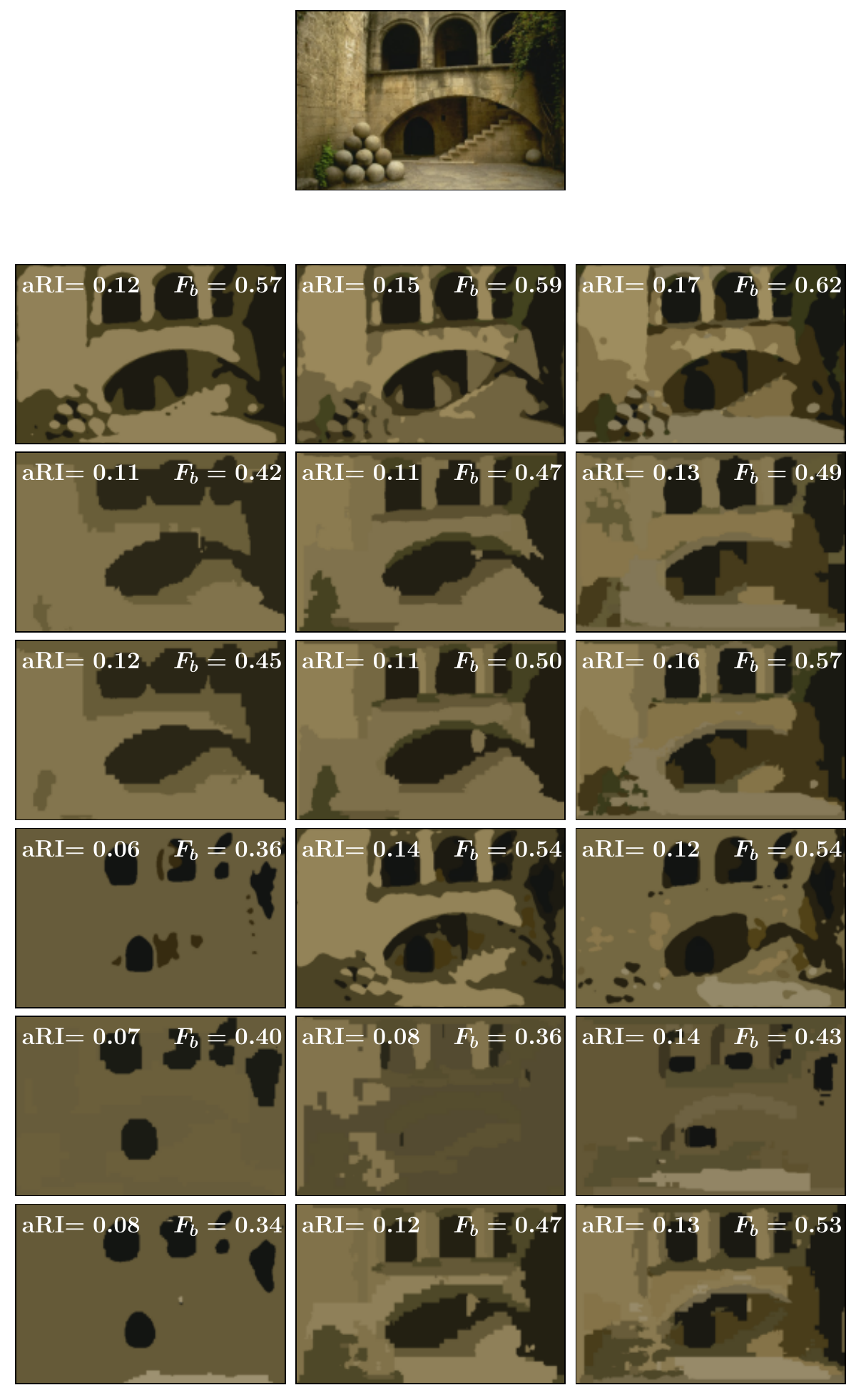}};
        \node at (3.5,-8.85){\includegraphics[width=7cm]{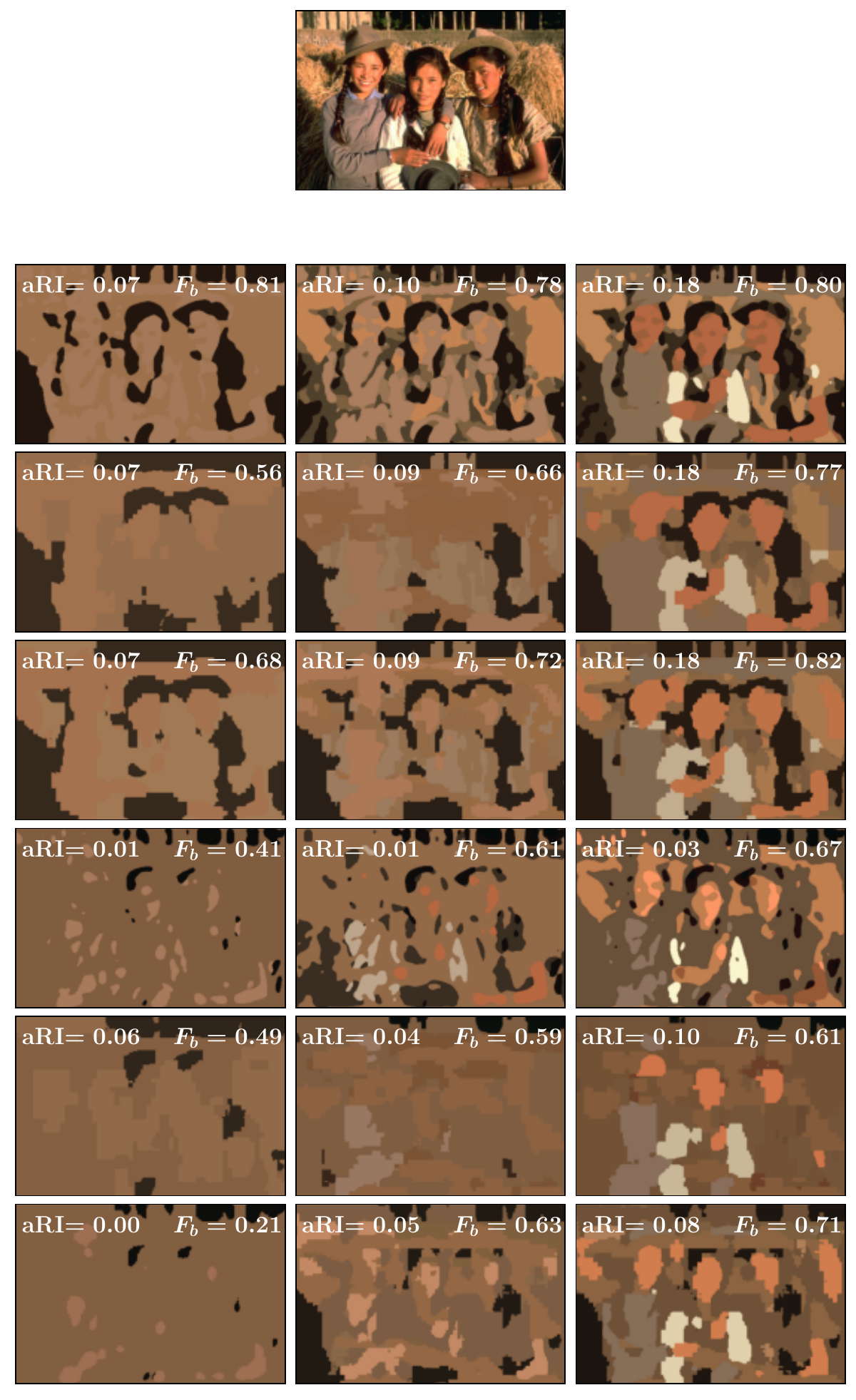}};
    \end{tikzpicture}
    \vspace{-2mm}
    \caption{Segmentation of natural images using models \textbf{a} ($1^{\text{st}}$ layer   and combined) and \textbf{b} on VGG 19 features with smoothing for Gaussian (sGMM) and Student-t mixture (sSMM). Top: high-aRI images. Bottom: low-aRI images. Figure is best seen in color.}
    \label{fig:ql-multiple-results}
\end{figure}

\begin{figure}
    \centering
    \begin{tikzpicture}
        \draw[step=1.0,white,thin] (-7.3,3.8) grid (7.3,-3.2);
        \node[text width=4.7cm] at (-3.7,3.4){sSMM-independent layers};
        \node at (-3.7,0.0){\includegraphics[width=7.4cm]{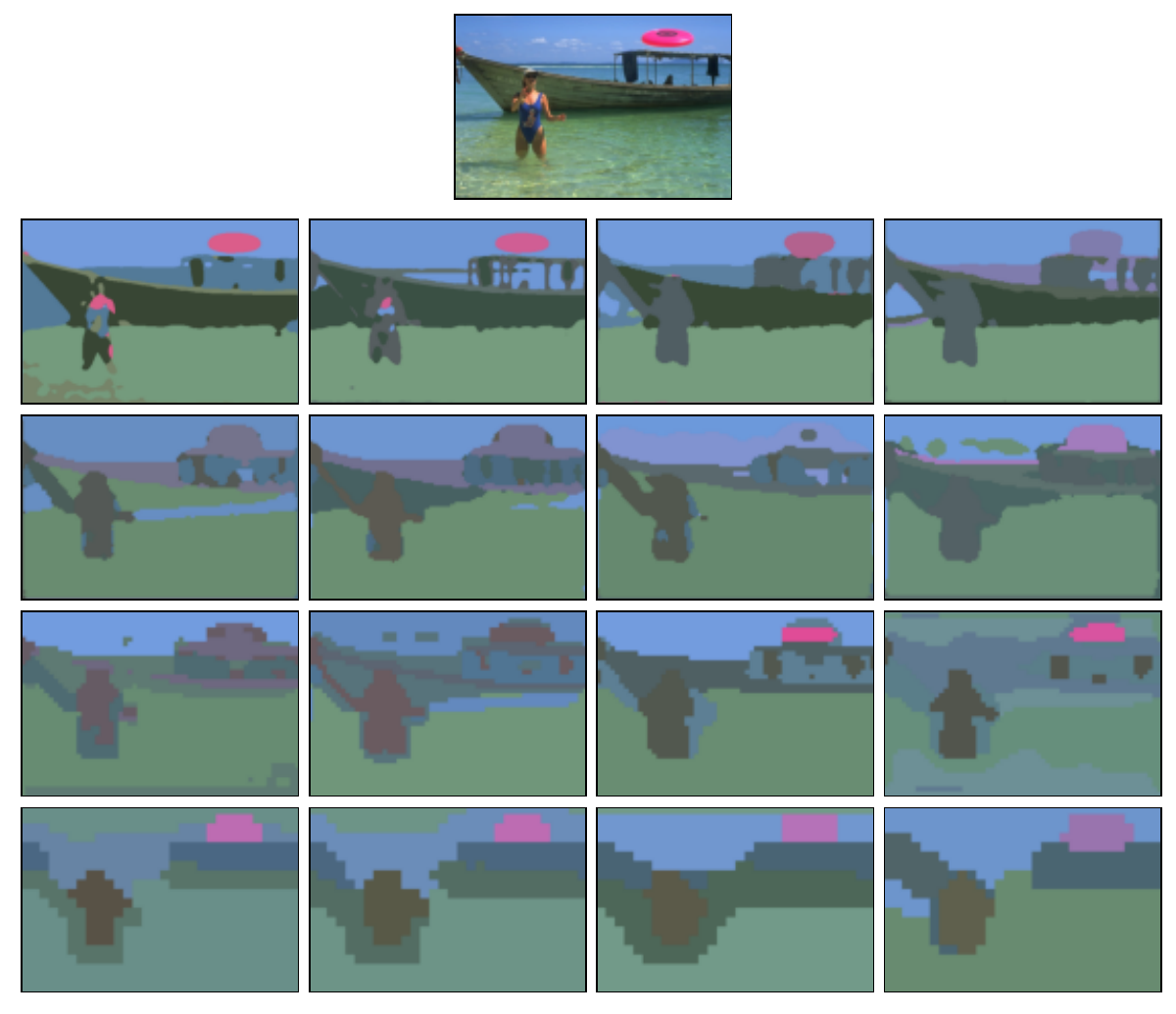}};
        \node[text width=1.6cm] at (3.65,3.4){sSMM-a};
        \node at (3.65,0.0){\includegraphics[width=7.4cm]{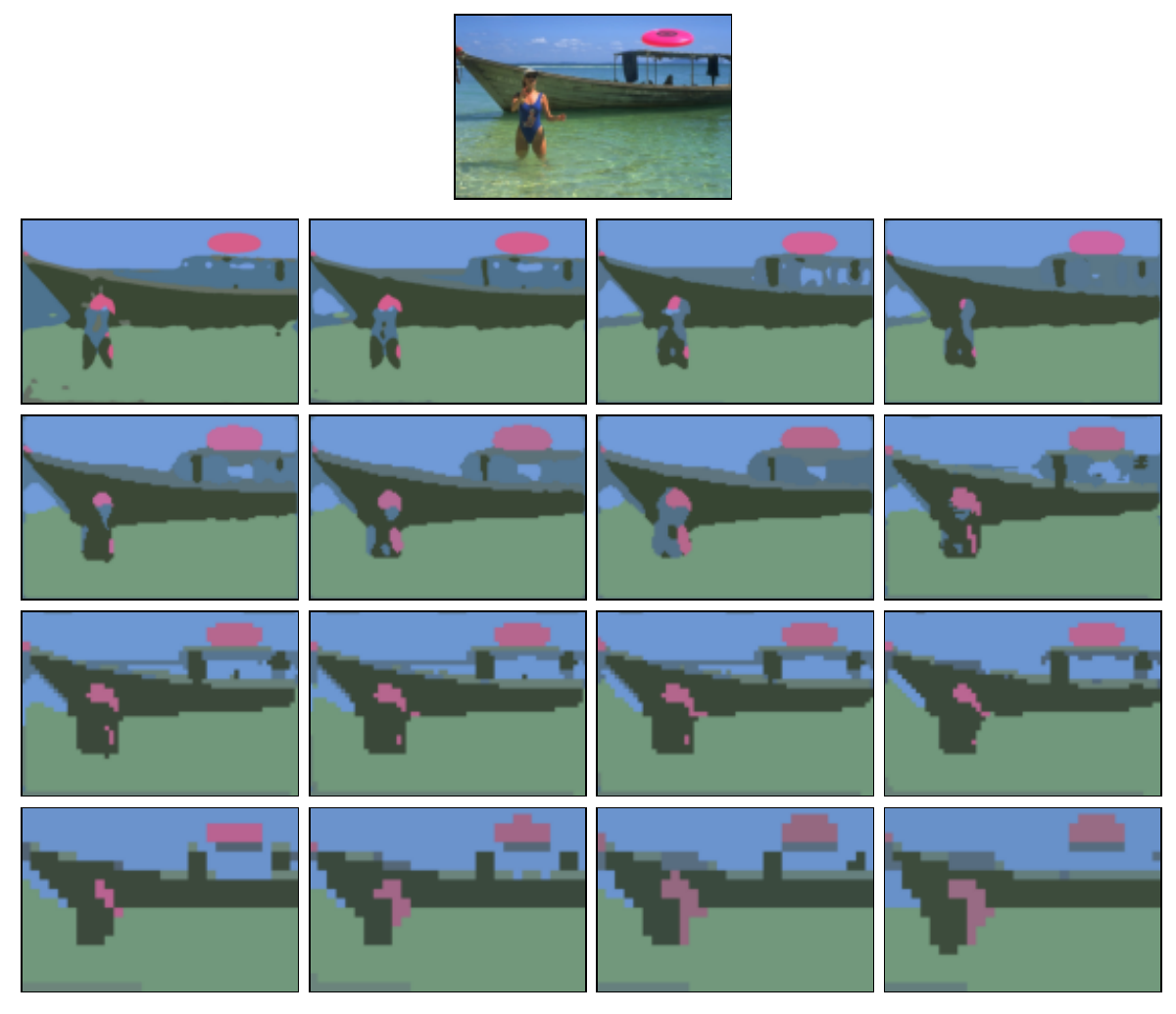}};
    \end{tikzpicture}
    \caption{Segmentation at all layers for $K=6$. Left: sSMM trained independently on each VGG layer. Right: sSMM-a. Figure is best seen in color.}
    \vspace{2mm}
    \label{fig:seg-maps-layers}
\end{figure}

\paragraph{Qualitative Results} We show in  Figure~\ref{fig:ql-multiple-results} four segmentation  examples, two with high aRI scores (top) and two with low aRI scores (bottom) for model \textbf{a} ($1^{\text{st}}$ layer and combined) and model \textbf{b}. \revisedB{These examples are selected to illustrate our main quantitative result: the overall superiority of SMM over GMM for all three models.} Similar to the segmentation based on RGB features, images with high scores tend to have sharper contours between segments and  more uniform textures within each segment. \revisedB{A second observation is that incorporating information from deeper layers (compare models \textbf{a} and \textbf{b} to model \textbf{a} $1^{\text{st}}$ layer) lower the spatial resolution of the segmentation maps.} For SMM, the segmentation maps provided by models \textbf{a} and \textbf{b} are hard to distinguish, indicating that  both combination methods (uncertainty weighting \vs{} product) lead to similar results for these four particular images. \revisedB{Nevertheless the negative correlation between scores and entropy (Figure~\ref{fig:score-vs-entropy}) suggests that model~\textbf{b} (\ie{} uncertainty weighting) is best across multiple images, as we observed in the quantitative results Figure~\ref{fig:qt-multiple-results}.} Finally, we illustrate in Figure~\ref{fig:seg-maps-layers} how our local combination of layers (model \textbf{a}) enforces consistency of segmentation maps across layers in comparison to when mixture models are trained on each layer independently.

%
\subsection{Discussion}

We have presented a new approach to unsupervised clustering via probabilistic mixture models, that exploits the topology of the dataset to improve clustering performance. We have shown that, similar to other related algorithms, our method can be understood by considering  the probability that each datapoint belongs to a latent class (the mixing probability) as a free parameter, and by introducing topological information as a strong regularization. Our main innovation consists in how we formulate the joint prior over mixing probabilities of all datapoints (\ie{} the regularizer): With our formulation, when optimizing the parameters via the EM algorithm, the update rule of the mixing probability of one datapoint is linear in the mixing probabilities of the other datapoints (Table~\ref{tab:update-comparison}). Furthermore, our method allows us to choose any desired linear combination of the mixing probabilities, and therefore can be adapted to arbitrary data topology.

We have demonstrated how the approach can be used to obtain spatial smoothing of clusters when the data have two spatial dimensions (\eg{} images; Figure~\ref{fig:graph-ours-mixture}), and uncertainty-weighted integration across multiple feature channels organized hierarchically (\eg{} across the layers of a DNN; Figure~\ref{fig:modified-mixture-multi}). We demonstrated on synthetic data that our method performs spatial smoothing and captures ground-truth clustering uncertainty accurately (Figures \ref{fig:artif-data} and \ref{fig:results-fit-multi-mixture}). In comparison to related methods ~\cite{nikou2010bayesian,sun2017location}, ours has the advantages of computational simplicity and flexibility, because it allows the user to choose any desired linear update rule, to combine information according to the data topology.

Our application to unsupervised segmentation of natural images, based on the pixels RGB values, demonstrates that smoothing is effective at removing spurious spatial discontinuities in segmentation maps, but also that, when doing so, it is important to adopt an observation model that is robust to outliers (\eg{} mixtures of Student-t rather than Gaussian distributions; Figures \ref{fig:qt-single-results} and \ref{fig:ql-single-results}). Our application to natural image segmentation based on VGG-19 demonstrates the flexibility of our approach to combine segmentation information across hierarchical layers. Because of the lower spatial resolution of deep layers, combining segmentation information across layers increases the resolution of the segmentation maps in deeper layers, while reducing it in superficial layers. In addition, our method produces segmentation maps that are consistent across all layers (Figure \ref{fig:ql-multiple-results}) and exploits higher-level information about objects (encoded implicitly by the deep layers of VGG-19) to correctly group areas that could otherwise be segmented based on low-level feature appearance, and to further improve noise reduction compared to spatial smoothing alone. 

Our analysis of quantitative performance on the BSD500 (Figures~\ref{fig:qt-single-results} and~\ref{fig:qt-multiple-results}) confirms the importance of using mixture components robust to outliers: the Student-t mixtures generally outperformed the Gaussian mixtures, particularly when using VGG19 features. The comparisons also highlight that information can be combined across the VGG19 hierarchy in different ways, leading to different outcomes. For the $F_b$ score (related to contour detection), a nearest-neighbor strategy performs better than a single prior map shared between all layers, because the former preserves higher spatial resolution; whereas for aRI (related to region overlap) there is not much difference. \revisedA{Overall, our models perform better than other comparable algorithms, including several standard approaches to unsupervised clustering (KMeans~\cite{steinhaus1956kmeans}, Birch~\cite{zhang1996birch}, MeanShift~\cite{comaniciu2002mean}) and the smoothing algorithms most closely related to our framework (DCM-SVFMM~\cite{nikou2010bayesian} and LDDP~\cite{sun2017location}). On the other hand, our models do not reach human-level performance, nor state-of-the-art performance set by more modern deep learning based algorithms for unsupervised image segmentation~\cite{maninis2018convolutional}}. \revisedB{Our algorithms performance could be improved by fine tuning certain implementation details, such as the dimensionality reduction of the feature space (e.g. \cite{bouveyron2014model}), and adopting methods to choose the best number of components $K$ per image.} \revisedB{However, maximizing performance on BSD and performing an exhaustive comparison with unsupervised segmentation methods is beyond the scope of this paper.}

\revisedB{As a model of biological vision, an important avenue} for improvement is suggested by the fact that image segmentation in human perception is influenced strongly by semantic information about the objects that are recognized in the image \cite{neri2017object}. Therefore, future extensions of our model for segmentation should include explicit object information, either directly, \eg{} using object classes predicted by VGG to bias the prior on mixing probabilities, or indirectly using human segmentation to refine the observation model of each cluster via semi-supervised learning. \revisedA{Besides performance on the standard benchmarks, the framework we have proposed has two main advantages over state-of-the-art algorithms~\cite{maninis2018convolutional}. First, it provides a probabilistic model, and therefore accounts for the uncertainty of segmentation, which will be crucial for real-life applications \eg{} with noisy sensors and complex scenes, as well as for models of biological visual segmentation. Although this cannot yet be quantified precisely with existing data sets, because segmentation uncertainty has not been measured, our analysis (Figure \ref{fig:score-vs-entropy}) indicates that the uncertainty captured by our algorithms can be exploited to identify candidate image regions where segmentation can be most improved. Second, different from existing deep learning based algorithms tailored for image segmentation, our model generalizes easily to other unsupervised learning problems and data topologies.}

The computational advantages of our approach come at the cost of introducing loops in the probabilistic graphical model, specifically between the class information and the prior on the class (Figure~\ref{fig:graph-ours-mixture}). In this paper, we have shown that this issue is mitigated by considering factor graphs theory and by observing that the loops are absent from the two subgraphs corresponding to the two steps of the EM algorithm.
Recurrent network models, that have been proposed recently to account for the fact that human image segmentation likely involves heavy recurrence and hierarchical feedback~\cite{linsley2019recurrent}, also rely on loopy architectures and are typically trained unrolling the model in time. Those models have been tested on specific tasks and benchmarks designed to highlight the importance of recurrence, but not on natural image segmentation databases. Although our approach is based on probabilistic graphical models and therefore is not directly comparable to recurrence and feedback in deterministic networks, our method to combine class information across datapoints effectively achieves similar goals. Additionally, our approach being fully probabilistic, it guarantees that the recurrent and feedback information are correctly weighted by their uncertainty, and the resulting segmentation maps are probabilistic. In future work, this could be exploited to study quantitatively recurrent processing in human perception and in neural processing in the visual cortex~\cite{vacher2018probabilistic,lee2003hierarchical}.

\section*{Acknowledgements}
We thank Pascal Mamassian for fruitful discussion. RCC is supported by NIH (NIH/CRCNS EY031166). JV is supported by ANR (ANR-19-NEUC-0003-01).

\bibliographystyle{apalike2}
\bibliography{references}

\appendix

\input{supplementary}

\end{document}

%% file: figs/figure-1.tex
\myfigurestar{
\subfloat[Standard mixture model.\label{fig:graph-std-mix}]{
\begin{tikzpicture}[line cap=round,line join=round,x=1cm,y=1cm,scale=1]
\draw[color=white, step=1] (-1.25,-3) grid (1.25,3);

\draw[thick] (0.00,1.8) node {$\mathbf{p}$};
\draw[thick,-T] (0,1.45)--(-0.6,0.95);
\draw[thick] (-0.6,0.6) circle (0.35cm) node {$C_n$};
\draw[thick,-T] (-0.6,0.25)--(-0.6,-0.25);
\draw[thick, fill=lightgray] (-0.6,-0.6) circle (0.35cm) node {$X_n$};
\draw[thick,-T] (0,1.45)--(0.6,0.95);
\draw[thick] (0.6,0.6) circle (0.35cm) node {$C_m$};
\draw[thick,-T] (0.6,0.25)--(0.6,-0.25);
\draw[thick, fill=lightgray] (0.6,-0.6) circle (0.35cm) node {$X_m$};
\draw[thick] (0,-1.8) node {$\mathbf{a}$};
\draw[thick,T-] (-0.6,-0.95)--(0,-1.45);
\draw[thick,T-] (0.6,-0.95)--(0,-1.45);

\end{tikzpicture}}\hfill
\subfloat[Location Dependent Dirichlet Process\label{fig:graph-lddp}]{
\begin{tikzpicture}[line cap=round,line join=round,x=1cm,y=1cm,scale=1]
\draw[color=white, step=1] (-1.5,-2) grid (1.5,4);

\draw[thick] (-1.1,3.3) circle (0.35cm) node {$\mathbf{F}_n$};
\draw[thick,-T] (-1.0,2.95)--(-0.705,2.12);
\draw[thick] (-0.6,1.8) circle (0.35cm) node {$\mathbf{P}_n$};
\draw[thick,-T] (-0.6,1.45)--(-0.6,0.95);
\draw[thick] (-0.6,0.6) circle (0.35cm) node {$C_n$};
\draw[thick,-T] (-0.6,0.25)--(-0.6,-0.25);
\draw[thick, fill=lightgray] (-0.6,-0.6) circle (0.35cm) node {$X_n$};
\draw[thick] (1.1,3.3) circle (0.35cm) node {$\mathbf{F}_m$};
\draw[thick,-T] (1.0,2.95)--(0.705,2.12);
\draw[thick] (0.6,1.8) circle (0.35cm) node {$\mathbf{P}_m$};
\draw[thick,-T] (0.6,1.45)--(0.6,0.95);
\draw[thick] (0.6,0.6) circle (0.35cm) node {$C_m$};
\draw[thick,-T] (0.6,0.25)--(0.6,-0.25);
\draw[thick, fill=lightgray] (0.6,-0.6) circle (0.35cm) node {$X_m$};
\draw[thick] (-0.75,3.3)--(0.75,3.3);
\draw[thick] (0,2.7) circle (0.35cm) node {$\mathbf{Q}$};
\draw[thick,-T] (0.2,2.4)--(0.45,2.1);
\draw[thick,-T] (-0.2,2.4)--(-0.45,2.1);
\draw[thick] (0,-1.8) node {$\mathbf{a}$};
\draw[thick,T-] (0.6,-0.95)--(0,-1.45);
\draw[thick,T-] (-0.6,-0.95)--(0,-1.45);

\end{tikzpicture}}\hfill
\subfloat[Spatially varying mixture models. \label{fig:graph-sv-mix}]{
\begin{tikzpicture}[line cap=round,line join=round,x=1cm,y=1cm,scale=1]
\draw[color=white, step=1] (-1.25,-2) grid (1.25,4);

\draw[thick] (-0.6,3.0) circle (0.35cm) node {$\mathbf{B}_n$};
\draw[thick,-T] (-0.6,2.65)--(-0.6,2.15);
\draw[thick] (-0.6,1.8) circle (0.35cm) node {$\mathbf{P}_n$};
\draw[thick,-T] (-0.6,1.45)--(-0.6,0.95);
\draw[thick] (-0.6,0.6) circle (0.35cm) node {$C_n$};
\draw[thick,-T] (-0.6,0.25)--(-0.6,-0.25);
\draw[thick, fill=lightgray] (-0.6,-0.6) circle (0.35cm) node {$X_n$};
\draw[thick] (0.6,3.0) circle (0.35cm) node {$\mathbf{B}_m$};
\draw[thick,-T] (0.6,2.65)--(0.6,2.15);
\draw[thick] (0.6,1.8) circle (0.35cm) node {$\mathbf{P}_m$};
\draw[thick,-T] (0.6,1.45)--(0.6,0.95);
\draw[thick] (0.6,0.6) circle (0.35cm) node {$C_m$};
\draw[thick,-T] (0.6,0.25)--(0.6,-0.25);
\draw[thick, fill=lightgray] (0.6,-0.6) circle (0.35cm) node {$X_m$};
\draw[thick,-] (-0.25,3)--(0.25,3);
\draw[thick] (0,-1.8) node {$\mathbf{a}$};
\draw[thick,T-] (0.6,-0.95)--(0,-1.45);
\draw[thick,T-] (-0.6,-0.95)--(0,-1.45);

\end{tikzpicture}}\hfill
\subfloat[Proposed mixture model. \label{fig:graph-ours-mixture}]{
\begin{tikzpicture}[line cap=round,line join=round,x=1cm,y=1cm,scale=1]
\draw[color=white, step=1] (-2.3,-3) grid (2.3,3);

\draw[thick] (-0.6,1.8) circle (0.35cm) node {$\mathbf{B}_n$};
\draw[thick,T-] (-1.45,1.8)--(-0.95,1.8);
\draw[thick] (-1.8,1.8) circle (0.35cm) node {$\mathbf{P}_n$};
\draw[thick,-T] (-1.55,1.55)--(-0.85,0.85);
\draw[thick] (-0.6,0.6) circle (0.35cm) node {$C_n$};
\draw[thick,T-] (-0.6,1.45)--(-0.6,0.95);
\draw[thick,-T] (-0.6,0.25)--(-0.6,-0.25);
\draw[thick, fill=lightgray] (-0.6,-0.6) circle (0.35cm) node {$X_n$};
\draw[thick] (0.6,1.8) circle (0.35cm) node {$\mathbf{B}_m$};
\draw[thick,T-] (1.45,1.8)--(0.95,1.8);
\draw[thick] (1.8,1.8) circle (0.35cm) node {$\mathbf{P}_m$};
\draw[thick,-T] (1.55,1.55)--(0.85,0.85);
\draw[thick] (0.6,0.6) circle (0.35cm) node {$C_m$};
\draw[thick,T-] (0.6,1.45)--(0.6,0.95);
\draw[thick,-T] (0.6,0.25)--(0.6,-0.25);
\draw[thick, fill=lightgray] (0.6,-0.6) circle (0.35cm) node {$X_m$};
\draw[thick,-T] (-0.35,0.85)--(0.35,1.55);
\draw[thick,-T] (0.35,0.85)--(-0.35,1.55);
\draw[thick] (0,-1.8) node {$\mathbf{a}$};
\draw[thick,T-] (0.6,-0.95)--(0,-1.45);
\draw[thick,T-] (-0.6,-0.95)--(0,-1.45);

\end{tikzpicture}}
}{Probabilistic graphical models corresponding to previously proposed mixture models (a-c) and our model (d) for two observed variables $X_n$ and $X_m$. Circles represent random variables (gray: observed; white: latent), uncircled letters represent parameters, directed and undirected edges represent probabilistic dependencies. The variables are described in the main text.}{fig:modified-mixture}

%% file: figs/figure-2.tex
\myfigurestar{
\subfloat[Initial factor graph\label{fig:factographall}]{
\begin{tikzpicture}[line cap=round,line join=round,x=1cm,y=1cm,scale=1]
\draw[color=white, step=1] (-2.25,-3) grid (2.25,3);

\draw[thick] (-0.8,2.4) circle (0.4cm) node (Bn) {$\mathbf{B}_n$};
\draw[thick] (0.8,2.4) circle (0.4cm) node (Bm) {$\mathbf{B}_m$};
\draw[thick] (-0.8,0.0) circle (0.4cm) node (Cn) {$C_n$};
\draw[thick] (0.8,0.0) circle (0.4cm) node (Cm) {$C_m$};
\draw[thick, fill=lightgray] (-0.8,-2.6) circle (0.4cm) node (Xn) {$X_n$};
\draw[thick, fill=lightgray] (0.8,-2.6) circle (0.4cm) node (Xm) {$X_m$};
\draw[thick] (0.0,-1.3) circle (0.4cm) node (a) {$\mathbf{a}$};
\draw[thick] (-1.8,-1.3) circle (0.4cm) node (Pn) {$\mathbf{P}_n$};
\draw[thick] (1.8,-1.3) circle (0.4cm) node (Pm) {$\mathbf{P}_m$};

\draw[L,T-] (Bn)--(Cn);
\draw[L,T-] (Bm)--(Cm);
\draw[L,-T] (Bn)--(Pn);
\draw[L,-T] (Bm)--(Pm);
\draw[L] (Cn)--(Xn);
\draw[L] (Cm)--(Xm);
\draw[thick,shorten > = 1.5mm] (Pn)--(a);
\draw[thick,shorten > = 1.5mm,shorten < = -0.5mm] (Pm)--(a);
\draw[thick,shorten < = 1.5mm,-Square] (a)--+(0,-1);
\draw[thick,shorten < = -1mm] (Cm)--+(-1.6,1.2);
\draw[thick,shorten < = -0.9mm] (Cn)--+(1.6,1.2);

\end{tikzpicture}}\hfill
\subfloat[E-factor graph\label{fig:factographE}]{
\begin{tikzpicture}[line cap=round,line join=round,x=1cm,y=1cm,scale=1]
\draw[color=white, step=1] (-2.25,-3) grid (2.25,3);

\draw[thick] (-0.8,2.4) circle (0.4cm) node (Bn) {$\mathbf{B}_n$};
\draw[thick] (0.8,2.4) circle (0.4cm) node (Bm) {$\mathbf{B}_m$};
\draw[thick] (-0.8,0.0) circle (0.4cm) node (Cn) {$C_n$};
\draw[thick] (0.8,0.0) circle (0.4cm) node (Cm) {$C_m$};
\draw[thick, fill=lightgray] (-0.8,-2.6) circle (0.4cm) node (Xn) {$X_n$};
\draw[thick, fill=lightgray] (0.8,-2.6) circle (0.4cm) node (Xm) {$X_m$};
\draw[thick, fill=lightgray] (0.0,-1.3) circle (0.4cm) node (a) {$\mathbf{a}$};
\draw[thick, fill=lightgray] (-1.8,-1.3) circle (0.4cm) node (Pn) {$\mathbf{P}_n$};
\draw[thick, fill=lightgray] (1.8,-1.3) circle (0.4cm) node (Pm) {$\mathbf{P}_m$};

\draw[L,T-] (Bn)--(Cn);
\draw[L,T-] (Bm)--(Cm);
\draw[L] (Cn)--(Xn);
\draw[L] (Cm)--(Xm);
\draw[thick,shorten > = 1.5mm] (Pn)--(a);
\draw[thick,shorten > = 1.5mm,shorten < = -0.5mm] (Pm)--(a);
\draw[thick,shorten < = 1.5mm,-Square] (a)--+(0,-1);
\draw[thick,shorten < = -1mm] (Cm)--+(-1.6,1.2);
\draw[thick,shorten < = -0.9mm] (Cn)--+(1.6,1.2);

\end{tikzpicture}}\hfill
\subfloat[M-factor graph\label{fig:factographM}]{
\begin{tikzpicture}[line cap=round,line join=round,x=1cm,y=1cm,scale=1]
\draw[color=white, step=1] (-2.25,-3) grid (2.25,3);

\draw[thick, fill=lightgray] (-0.8,2.4) circle (0.4cm) node (Bn) {$\mathbf{B}_n$};
\draw[thick, fill=lightgray] (0.8,2.4) circle (0.4cm) node (Bm) {$\mathbf{B}_m$};
\draw[thick, fill=lightgray] (-0.8,0.0) circle (0.4cm) node (Cn) {$C_n$};
\draw[thick, fill=lightgray] (0.8,0.0) circle (0.4cm) node (Cm) {$C_m$};
\draw[thick, fill=lightgray] (-0.8,-2.6) circle (0.4cm) node (Xn) {$X_n$};
\draw[thick, fill=lightgray] (0.8,-2.6) circle (0.4cm) node (Xm) {$X_m$};
\draw[thick] (0.0,-1.3) circle (0.4cm) node (a) {$\mathbf{a}$};
\draw[thick] (-1.8,-1.3) circle (0.4cm) node (Pn) {$\mathbf{P}_n$};
\draw[thick] (1.8,-1.3) circle (0.4cm) node (Pm) {$\mathbf{P}_m$};

\draw[L,-T] (Bn)--(Pn);
\draw[L,-T] (Bm)--(Pm);
\draw[L] (Cn)--(Xn);
\draw[L] (Cm)--(Xm);
\draw[thick,shorten < = 6mm,,shorten > = 1.5mm] (Pn)--(a);
\draw[thick,shorten > = 1.5mm,shorten < = 6mm] (Pm)--(a);
\draw[thick,shorten < = 1.5mm,-Square] (a)--+(0,-1);

\end{tikzpicture}}
}{Factor graphs associated with our model, the factorization defined in~\eqref{eq:joint-distribution_alldata} and with each step of the EM algorithm. The graphs connect each factor node (black squares) to the corresponding variables (gray circles: observed variable; white circles: latent variables) as given in Equations (\ref{eq:joint-distribution_alldata}-\ref{eq:factor-a}). Arrows indicate conditional probabilities.}{fig:factorgraphs}

%% file: figs/figure-3.tex
\tikzset{pics/graph/.style args={#1}{
code={
    \draw[thick] (-0.6,1.8) circle (0.35cm) node {$\mathbf{B}^{{\scriptscriptstyle #1}}_n$};
    \draw[thick,T-] (-1.45,1.8)--(-0.95,1.8);
    \draw[thick] (-1.8,1.8) circle (0.35cm) node {$\mathbf{P}^{{\scriptscriptstyle #1}}_n$};
    \draw[thick,-T] (-1.55,1.55)--(-0.85,0.85);
    \draw[thick] (-0.6,0.6) circle (0.35cm) node {$C^{{\scriptscriptstyle #1}}_n$};
    \draw[thick,T-] (-0.6,1.45)--(-0.6,0.95);
    \draw[thick,-T] (-0.6,0.25)--(-0.6,-0.25);
    \draw[thick, fill=lightgray] (-0.6,-0.6) circle (0.35cm) node {$X^{{\scriptscriptstyle #1}}_n$};
    \draw[thick] (0.6,1.8) circle (0.35cm) node {$\mathbf{B}^{{\scriptscriptstyle #1}}_m$};
    \draw[thick,T-] (1.45,1.8)--(0.95,1.8);
    \draw[thick] (1.8,1.8) circle (0.35cm) node {$\mathbf{P}^{{\scriptscriptstyle #1}}_m$};
    \draw[thick,-T] (1.55,1.55)--(0.85,0.85);
    \draw[thick] (0.6,0.6) circle (0.35cm) node {$C^{{\scriptscriptstyle #1}}_m$};
    \draw[thick,T-] (0.6,1.45)--(0.6,0.95);
    \draw[thick,-T] (0.6,0.25)--(0.6,-0.25);
    \draw[thick, fill=lightgray] (0.6,-0.6) circle (0.35cm) node {$X^{{\scriptscriptstyle #1}}_m$};
    \draw[thick,-T] (-0.35,0.85)--(0.35,1.55);
    \draw[thick,-T] (0.35,0.85)--(-0.35,1.55);
    \draw[thick] (0,-1.8) node {$\mathbf{a}^{{\scriptscriptstyle #1}}$};
    \draw[thick,T-] (0.6,-0.95)--(0,-1.45);
    \draw[thick,T-] (-0.6,-0.95)--(0,-1.45);
}}}
\myfigurestar{
\subfloat{
\begin{tikzpicture}[line cap=round,line join=round,x=1cm,y=1cm,scale=1]
\draw[color=white, step=1] (-7.5,-2.1) grid (7.5,2.3);

\pic at (-5,0) {graph={(1)}};
\pic at (0,0) {graph={(2)}};
\pic at (5,0) {graph={(3)}};
\draw[line width=1.05pt, white] (-0.6,1.8) circle (0.35cm);
\draw[thick, orange] (-0.6,1.8) circle (0.35cm);
\draw[orange, thick] (-5,0.6) ellipse (15mm and 5mm);
\draw[orange, thick,-T] (-3.5,0.6)--(-0.85,1.55);
\draw[orange, thick] (5,0.6) ellipse (15mm and 5mm);
\draw[orange, thick,-T] (3.5,0.6)--(-0.35,1.55);

\end{tikzpicture}}
}{Mixture model for two observed variables $X_n$ and $X_m$ and 3 layers. Our model enables the combination of the 3 mixtures models, each one associated to a layer. In this graph, we focus on the conditional dependencies of $B_n^{(2)}$ which are determined by $K \times  H$ linear functions $(u_{n,k}^{(h)})_{k,h}$.}{fig:modified-mixture-multi}

%% file: supplementary.tex
\section{Supplementary Figures}
\label{sec:supp-fig}

\begin{figure}[H]
    \centering
    \includegraphics[width=6cm]{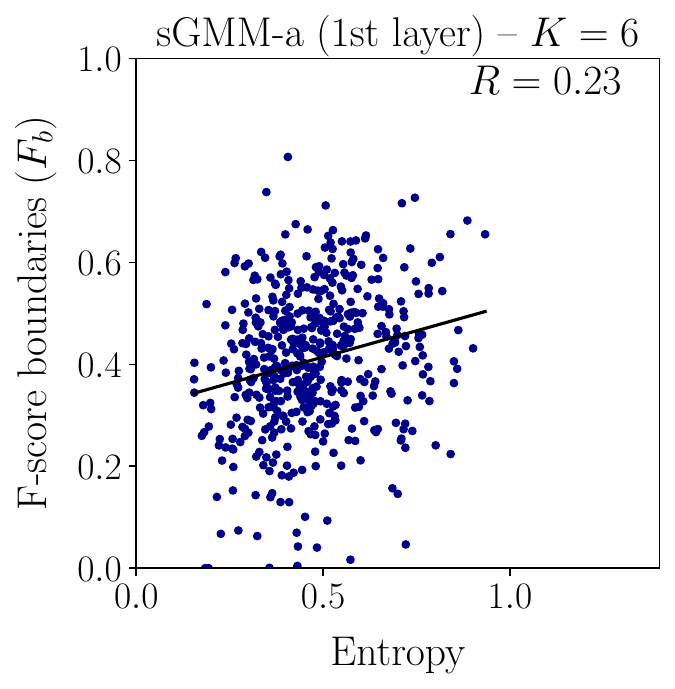}
    \includegraphics[width=6cm]{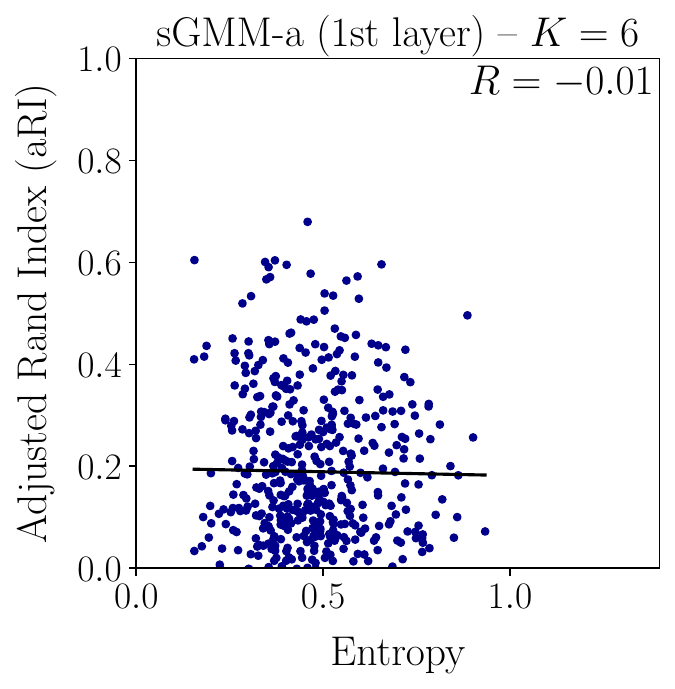}
    \caption{Positive or null correlation between both scores and entropy model \textbf{a}/1st layer when using Gaussian mixtures.}
    \label{supp-fig:score-vs-entropy}
\end{figure}



\section{Proof of propositions \ref{prop:gen-update-rule} and \ref{prop:gen-update-rule-ext}} 
\label{asec:proof}

\begin{proof}[Proof of Proposition \ref{prop:gen-update-rule}]
Using the Dirichlet prior \eqref{eq:Dirichlet} of the main paper, the completed log-posterior writes
\eq{
\ell\left( \boldsymbol{\theta}; (x_n, C_n,\mathbf{B}_n)_{n} \right) =    \sum_{n=1}^{N} \sum_{k=1}^{K}  ({B}_{n,k}-1 + \one_k(C_n))\ln\left( p_{n,k}  \right) + W((x_n, C_n, \mathbf{B}_n)_{n};\boldsymbol{\alpha}),
}
where $W$ is the function that gathers all the terms of $\ell$ that do not depend on $p_{n,k}$.
Knowing the previous parameter estimate $\boldsymbol{\theta}^{(t)}$, the E-step consists in taking the conditional expectation of the log-posterior $\ell$ which is
\eq{
Q(\boldsymbol{\theta}; \boldsymbol{\theta}^{(t)}, (x_n)_n) =  \sum_{n=1}^{N} \sum_{k=1}^{K}  \EE\left({B}_{n,k}-1 + \one_k(C_n) \vert (x_n)_n,\boldsymbol{\theta}^{(t)}\right) \ln\left( p_{n,k}  \right) + w((x_n)_{n};\boldsymbol{\alpha},\boldsymbol{\theta}^{(t)}),
}
where $w((x_n)_{n};\boldsymbol{\alpha},\boldsymbol{\theta}^{(t)}) = \EE\left( W((x_n, C_n, \mathbf{B}_n)_{n};\boldsymbol{\alpha}) \vert (x_n)_n,\boldsymbol{\theta}^{(t)}\right)$
and 

\eqAL{
\EE\left({B}_{n,k}-1 + \one_k(C_n) \vert (x_n)_n,\boldsymbol{\theta}^{(t)}\right) & = \EE\left( u_{n,k}(\one_k(C_\cdot)) \vert (x_n)_n,\boldsymbol{\theta}^{(t)}\right) \\
& = u_{n,k}\left( \EE\left( \one_k(C_\cdot) \vert (x_n)_n,\boldsymbol{\theta}^{(t)} \right) \right) =  u_{n,k}\left( \tau_{\cdot,k}^{(t)} \right),
}
where $\tau_{n,k}^{(t)}=\cpr{C_n}{X_n,\boldsymbol{\Theta}}(k|x_n,\boldsymbol{\theta}^{(t)})$. 
Then, the M-step consists in maximizing the expected log-posterior $Q$ with respect to $\boldsymbol{\theta}=(\mathbf{p}_n,\boldsymbol{\alpha})$. We only consider optimization with respect to $\mathbf{p}_n$ which is independent from the optimization with respect to $\boldsymbol{\alpha}$. To obtain the update rule for $\mathbf{p}_n$ with first add the Lagrange multiplier associated to the constraint $ \sum_k p_{n,k} = 1$ and compute the partial derivative with respect to $p_{n,k}$. Therefore,
\eq{
\forall (n,k) \in \{1,\dots,N\}\times \{1,\dots,K\},  \quad \frac{u_{n,k}\left(\tau_{\cdot,k}^{(t)}\right)}{p_{n,k}} + \lambda_n = 0,
}
which leads to the update rule~\eqref{eq:gen-update-rule} by setting $\lambda_n$ such that $ \sum_k p_{n,k} = 1$.
\end{proof}

\begin{proof}[Proof of Proposition \ref{prop:gen-update-rule-ext}]
The proof is similar to the proof of Proposition~\ref{prop:gen-update-rule} and starts by writing the log-posterior of each layers and then taking the conditional expectation knowing all the other features and previous parameter estimations at all layers
\eqAL{
\EE\left({B}_{n,k}^{(h)}-1 + \one_k(C_n^{(h)}) \Big\vert (x_n)_n,\boldsymbol{\theta}^{(t)}\right) & = \EE\left( u_{n,k}^{(h)}\left(\one_k(C_\cdot^{(1)}),\dots,\one_k(C_\cdot^{(H)})\right) \Big\vert \left((x_n^{(h)})_{n},\boldsymbol{\theta}^{(t,h)}\right)_h \right) \\
& = u_{n,k}^{(h)}\left( \EE\left( \left(\one_k(C_\cdot^{(1)}),\dots,\one_k(C_\cdot^{(H)})\right) \Big\vert \left((x_n^{(h)})_{n},\boldsymbol{\theta}^{(t,h)}\right)_h \right)  \right)\\
& = u_{n,k}^{(h)}(\tau_{\cdot,k}^{(t,1)},\dots,\tau_{\cdot,k}^{(t,H)}).
}
We conclude by using Lagrange multipliers as in the proof of Proposition~\ref{prop:gen-update-rule}.
\end{proof}

\section{Proof of propositions \ref{prop:EM-increases-logpost}} 
\label{asec:increase}

\begin{proof}[Proof that the EM algorithm increases the log-posterior of mixture models]

First, let us recall why each iteration of the EM algorithm applied to a general probabilistic mixture models as presented in the introduction increases the (incomplete) log-likelihood.
At iteration $t+1$, the E-step of the algorithm consists in computing 
\eq{Q(\boldsymbol{\theta}; \boldsymbol{\theta}^{(t)}, (x_n)_n)= \mathbb{E}_{C \vert X,\boldsymbol{\theta}^{(t)}} \left(\ln \ \mathbb{P}_{(X_n)_n,(C_n)_n \vert \boldsymbol{\theta}}( (x_n)_n, (C_n)_n \vert \boldsymbol{\theta}) \right),}
while the M-step maximizes $Q(\boldsymbol{\theta}; \boldsymbol{\theta}^{(t)}, (x_n)_n) \text{ with respect to }\boldsymbol{\theta}.$ For clarity purposes, we drop probability subscripts in the following.
First, using the chain rule, we have $\mathbb{P}((x_n)_n,(C_n)_n \vert \boldsymbol{\theta}) = \mathbb{P}((C_n)_n \vert (x_n)_n, \boldsymbol{\theta}) \mathbb{P}( (x_n)_n \vert \boldsymbol{\theta})$, so for all $\boldsymbol{\theta}$,
\eq{\ln (\mathbb{P}( (x_n)_n \vert \boldsymbol{\theta})) = \ln( \mathbb{P}((x_n)_n,(C_n)_n \vert \boldsymbol{\theta})) - \ln (\mathbb{P}((C_n)_n \vert (x_n)_n, \boldsymbol{\theta}) ).}
Then,
\eqAL{\mathbb{E}_{C \vert X,\boldsymbol{\theta}^{(t)}} (\ln (\mathbb{P}( (x_n)_n \vert \boldsymbol{\theta}))) &= \mathbb{E}_{C \vert X,\boldsymbol{\theta}^{(t)}} (\ln( \mathbb{P}((x_n)_n,(C_n)_n \vert \boldsymbol{\theta}))) - \mathbb{E}_{C \vert X,\boldsymbol{\theta}^{(t)}} (\ln (\mathbb{P}((C_n)_n \vert (x_n)_n, \boldsymbol{\theta}) )\\
\ln (\mathbb{P}( (x_n)_n \vert \boldsymbol{\theta}))  &= Q(\boldsymbol{\theta}; \boldsymbol{\theta}^{(t)}, (x_n)_n) + H\left((C_n)_n \vert (x_n)_n, \boldsymbol{\theta}^{(t)}, (C_n)_n \vert (x_n)_n, \boldsymbol{\theta}\right),}
where $H(p,q)$ is the cross-entropy of the probability distributions $p$ and $q$. The previous equation holds for all parameters $\boldsymbol{\theta}$ so, by subtracting the same equation for $\boldsymbol{\theta} = \boldsymbol{\theta}^{(t)}$, we obtain 
\eqAL{\ln (\mathbb{P}( (x_n)_n \vert \boldsymbol{\theta})) &- \ln (\mathbb{P}( (x_n)_n \vert \boldsymbol{\theta}^{(t)})) = Q(\boldsymbol{\theta}; \boldsymbol{\theta}^{(t)}, (x_n)_n) - Q(\boldsymbol{\theta}^{(t)}; \boldsymbol{\theta}^{(t)}, (x_n)_n) + \\
&H\left((C_n)_n \vert (x_n)_n, \boldsymbol{\theta}^{(t)}, (C_n)_n \vert (x_n)_n, \boldsymbol{\theta}\right) - H\left((C_n)_n \vert (x_n)_n, \boldsymbol{\theta}^{(t)}, (C_n)_n \vert (x_n)_n, \boldsymbol{\theta}^{(t)}\right).}
The Gibbs' inequality \cite{murphy2012machine} states that for all probability distributions $p$ and $q$, $H(p,q) \geq H(p,p)$, so
\eq{
\ln (\mathbb{P}( (x_n)_n \vert \boldsymbol{\theta})) - \ln (\mathbb{P}( (x_n)_n \vert \boldsymbol{\theta}^{(t)})) \geq Q(\boldsymbol{\theta}; \boldsymbol{\theta}^{(t)}, (x_n)_n) - Q(\boldsymbol{\theta}^{(t)}; \boldsymbol{\theta}^{(t)}, (x_n)_n).}
At iteration $t+1$; the M-step consists in defining $\boldsymbol{\theta}^{(t+1)}$ such that it maximizes $Q(\boldsymbol{\theta}; \boldsymbol{\theta}^{(t)}, (x_n)_n)$. Therefore,
\eq{
\ln (\mathbb{P}( (x_n)_n \vert \boldsymbol{\theta}^{(t)})) - \ln (\mathbb{P}( (x_n)_n \vert \boldsymbol{\theta}^{t})) \geq 0.}
\end{proof}

\begin{proof}[Proof that EM algorithm increases the log-posterior of our models]
Similarly, in our framework, each iteration the EM-algorithm applied to the complete log-posterior increases the incomplete log-posterior. $\boldsymbol{\theta}$ denotes the model parameters $\boldsymbol{a}, (\boldsymbol{p}_n)_n$. At iteration $t+1$, the E-step computes
\eqAL{Q(\boldsymbol{\theta};& \boldsymbol{\theta}^{(t)}, (x_n)_n)= \mathbb{E}_{C,\mathbf{B} \vert X,\boldsymbol{\theta}^{(t)}} \left( \ln \cpr{\boldsymbol{\theta}}{X,C,\mathbf{B}}(\boldsymbol{\theta}\vert (x_n)_n,(C_n)_n,(\mathbf{B}_n)_n)  \right)\\
&=\mathbb{E}_{C,\mathbf{B} \vert X,\boldsymbol{\theta}^{(t)}} \left(\ln \left( \mathbb{P}_{X,C,\mathbf{B},\boldsymbol{\theta}}( (x_n)_n, (C_n)_n, (\mathbf{B}_n)_n, \boldsymbol{\theta})\right) - \ln \left( \mathbb{P}_{X,C,\mathbf{B}}( (x_n)_n, (C_n)_n, (\mathbf{B}_n)_n)\right) \right),}
and the M-step maximizes $Q(\boldsymbol{\theta}; \boldsymbol{\theta}^{(t)}, (x_n)_n)$ with respect to $\boldsymbol{\theta}$. Once again, we drop the probability subscripts for clarity purposes and using the chain rule, we have
\eq{
\pr{}( (x_n)_n, (C_n)_n, (\mathbf{B}_n)_n, \boldsymbol{\theta}) = \pr{}( (C_n)_n, (\mathbf{B}_n)_n \vert  (x_n)_n,\boldsymbol{\theta}) \pr{}(\boldsymbol{\theta} \vert (x_n)_n) \pr{}((x_n)_n),}
so for all $\boldsymbol{\theta}$,
\eq{
\ln \pr{}(\boldsymbol{\theta} \vert (x_n)_n) = \ln \left( \pr{}( (x_n)_n, (C_n)_n, (\mathbf{B}_n)_n, \boldsymbol{\theta})\right) - \ln \left(\pr{}( (C_n)_n, (\mathbf{B}_n)_n \vert  (x_n)_n,\boldsymbol{\theta})\right) - \ln \left( \pr{}((x_n)_n)\right).}
Then,
\eqAL{
\mathbb{E}_{C,\mathbf{B} \vert X,\boldsymbol{\theta}^{(t)}} (\ln &\pr{}(\boldsymbol{\theta} \vert (x_n)_n)) = \mathbb{E}_{C,\mathbf{B} \vert X,\boldsymbol{\theta}^{(t)}} (\ln \pr{}( (x_n)_n, (C_n)_n, (\mathbf{B}_n)_n, \boldsymbol{\theta}))\\
&- \mathbb{E}_{C,\mathbf{B} \vert X,\boldsymbol{\theta}^{(t)}} (\ln \pr{}( (C_n)_n, (\mathbf{B}_n)_n \vert  (x_n)_n,\boldsymbol{\theta}) ) - \mathbb{E}_{C,\mathbf{B} \vert X,\boldsymbol{\theta}^{(t)}} (\ln \pr{}((x_n)_n))}
\eqAL{\ln \pr{}(\boldsymbol{\theta} \vert (x_n)_n)  = & Q(\boldsymbol{\theta}; \boldsymbol{\theta}^{(t)}, (x_n)_n) + \mathbb{E}_{C,\mathbf{B} \vert X,\boldsymbol{\theta}^{(t)}} \left( \ln \left( \mathbb{P}_{X,C,\mathbf{B}}( (x_n)_n, (C_n)_n, (\mathbf{B}_n)_n)\right) \right)\\
&+ H((C_n)_n, (\mathbf{B}_n)_n  \vert (x_n)_n, \boldsymbol{\theta}^{(t)}, (C_n)_n, (\mathbf{B}_n)_n \vert (x_n)_n, \boldsymbol{\theta}) - \ln \pr{}((x_n)_n),}
As before, by subtracting the same equation for $\boldsymbol{\theta} = \boldsymbol{\theta}^{(t)}$ and thanks to the Gibbs' inequality, we obtain 
\eq{\ln \pr{}(\boldsymbol{\theta} \vert (x_n)_n)  - \ln \pr{}(\boldsymbol{\theta}^{(t)}\vert (x_n)_n)  \geq Q(\boldsymbol{\theta}; \boldsymbol{\theta}^{(t)}, (x_n)_n) - Q(\boldsymbol{\theta}^{(t)}; \boldsymbol{\theta}^{(t)}, (x_n)_n).}
The M-step at iteration $t+1$ consists in defining $\boldsymbol{\theta}^{(t+1)}$ such that it maximizes $Q(\boldsymbol{\theta}; \boldsymbol{\theta}^{(t)}, (x_n)_n)$. Therefore,
\eq{\label{eq:last}
\ln \pr{}(\boldsymbol{\theta}^{(t+1)} \vert (x_n)_n) - \ln \pr{}(\boldsymbol{\theta}^{(t)} \vert (x_n)_n) \geq 0.
}
Note that the proof is similar for our model combining mixture models. 
\end{proof}